\documentclass[10pt,twocolumn,letterpaper]{article}

\usepackage{iccv}
\usepackage{times}
\usepackage{epsfig}
\usepackage{graphicx}
\usepackage{amsmath}
\usepackage{amssymb}
\usepackage{pifont} 
\usepackage{booktabs}
\usepackage{multirow, makecell}   
\usepackage{color, colortbl}   
\usepackage[table,xcdraw]{xcolor}
\usepackage{algorithm, algorithmicx, algpseudocode}

\usepackage[accsupp]{axessibility} 

\newcommand{\cmark}{\ding{51}}
\newcommand{\xmark}{\ding{55}}
\definecolor{mblue}{rgb}{ 0,  .439,  .753}

\usepackage[pagebackref=true,breaklinks=true,letterpaper=true,colorlinks,bookmarks=false]{hyperref}

\iccvfinalcopy 

\def\iccvPaperID{6788} 

\ificcvfinal\fi

\begin{document}

\title{Label Shift Adapter for Test-Time Adaptation under Covariate and Label Shifts}

\newcommand*{\affmark}[1][*]{\textsuperscript{#1}}

\def\@fnsymbol#1{\ensuremath{\ifcase#1\or \dagger\or \ddagger\or
   \mathsection\or \mathparagraph\or \|\or **\or \dagger\dagger
   \or \ddagger\ddagger \else\@ctrerr\fi}}
    \makeatother

\author{
Sunghyun Park\affmark[1,2] \; Seunghan Yang\affmark[1]\; Jaegul Choo\affmark[2] \; Sungrack Yun\affmark[1]\\
\affmark[1]Qualcomm AI Research\thanks{Qualcomm AI Research is an initiative of Qualcomm Technologies, Inc.} \; \affmark[2]KAIST\\
\texttt{\footnotesize\affmark[1]\{sunpar, seunghan, sungrack\}@qti.qualcomm.com} \; 
\texttt{\footnotesize\affmark[2]\{psh01087, jchoo\}@kaist.ac.kr }\ \\
}


\maketitle
\ificcvfinal\thispagestyle{empty}\fi

\begin{abstract}
Test-time adaptation (TTA) aims to adapt a pre-trained model to the target domain in a batch-by-batch manner during inference.
While label distributions often exhibit imbalances in real-world scenarios, most previous TTA approaches typically assume that both source and target domain datasets have balanced label distribution.
Due to the fact that certain classes appear more frequently in certain domains (e.g., buildings in cities, trees in forests), it is natural that the label distribution shifts as the domain changes.
However, we discover that the majority of existing TTA methods fail to address the coexistence of covariate and label shifts.
To tackle this challenge, we propose a novel label shift adapter that can be incorporated into existing TTA approaches to deal with label shifts during the TTA process effectively.
Specifically, we estimate the label distribution of the target domain to feed it into the label shift adapter.
Subsequently, the label shift adapter produces optimal parameters for target label distribution.
By predicting only the parameters for a part of the pre-trained source model, our approach is computationally efficient and can be easily applied, regardless of the model architectures.
Through extensive experiments, we demonstrate that integrating our strategy with TTA approaches leads to substantial performance improvements under the joint presence of label and covariate shifts.
\end{abstract}

\section{Introduction}

Despite the recent remarkable improvement of deep neural networks in various applications, the models still suffer from distribution shifts between source distribution and target distribution.
One type of distribution shift, known as \emph{covariate shift}, occurs when the target distribution $p_t(x)$ differs from the source distribution $p_s(x)$.
In autonomous driving, for instance, models may degrade significantly during testing due to ambient factors such as weather and location.
To design the models robust to covariate shifts, unsupervised domain adaptation literature~\cite{ganin2015unsupervised,ganin2016domain,liang2020we,kurmi2021domain} has explored the transfer of knowledge learned from labeled source data to unlabeled target data.

\begin{figure}[t!]
    \centering
    \includegraphics[width=\linewidth]{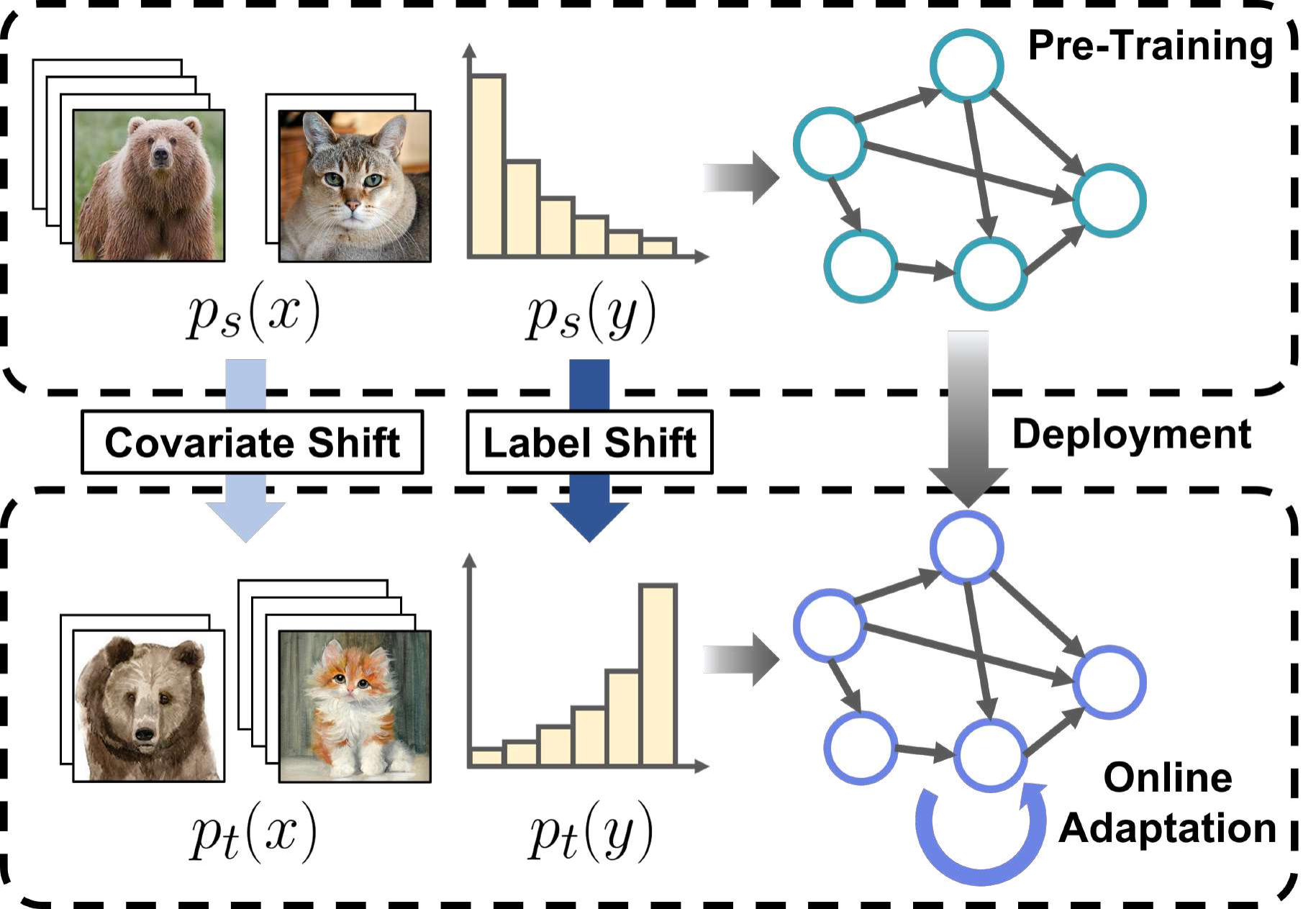}
    \caption{We consider the test-time adaptation scenario, when covariate and label shifts occur simultaneously. After deploying the pre-trained model, the model is adapted to the target domain. However, existing methods often suffer from the coexistence of covariate and label shifts. Employing our method enables online adaptation under shifted target label distributions.}
    \vspace{-0.6cm}
    \label{fig:task}
\end{figure}

To be more practical in real-world scenarios, test-time adaptation (TTA) algorithms~\cite{wang2020tent} have emerged to enhance practicality in real-world scenarios by adapting deep neural networks to the target domain during inference.
Specifically, TTA approaches optimize the model parameters batch-by-batch using unlabeled test data, avoiding additional labeling costs.
Previous TTA studies have mitigated the performance degradation caused by covariate shift by enhancing normalization statistics~\cite{schneider2020improving,gong2022note,lim2023ttn}, optimizing model parameters with entropy minimization~\cite{wang2020tent,niu2022efficient}, or utilizing pseudo labels~\cite{lee2013pseudo}.

\begin{figure}[t!]
    \centering
    \includegraphics[width=\linewidth]{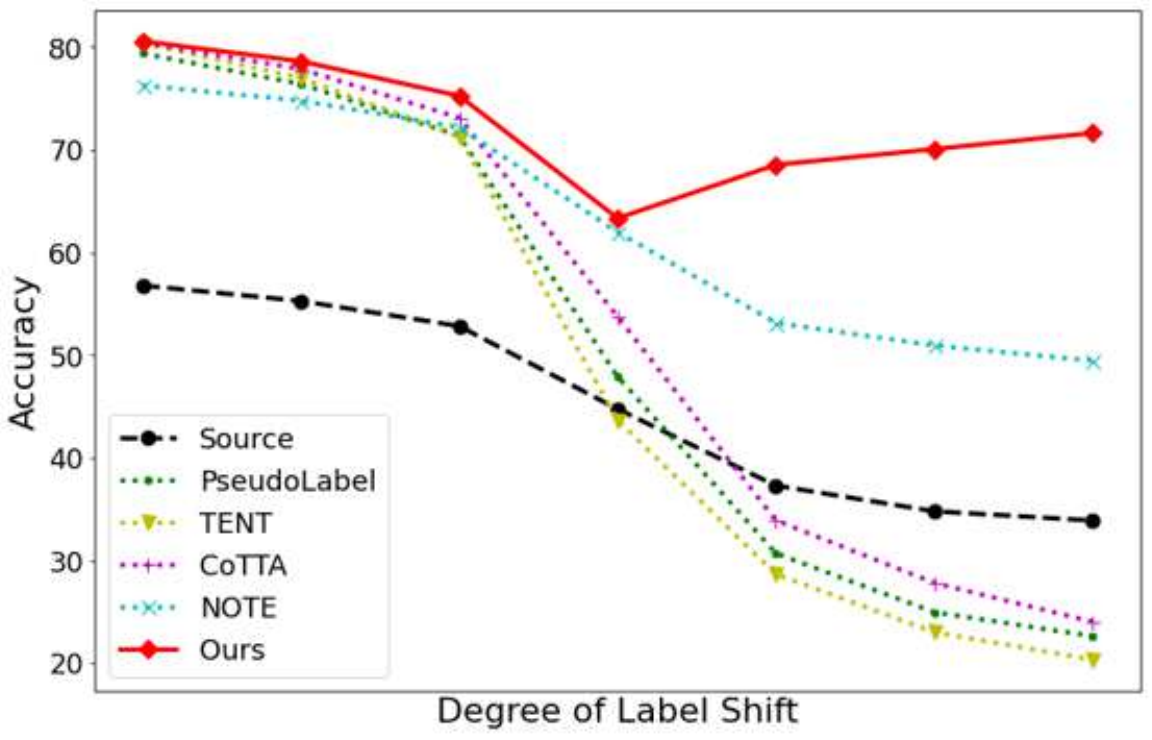}
    \caption{Plots of accuracy with different degrees of label shift in CIFAR-10-C~\cite{hendrycks2018benchmarking}. Although we train source model using balanced softmax~\cite{ren2020balanced}, most TTA baselines show lower performance than source model, when label shift is severe. 
    On the other hand, our method is more robust to label shifts than other TTA methods.}
    \vspace{-0.5cm}
    \label{fig:motivation}
\end{figure}

Although the natural data encountered in practice often exhibits long-tailed label distribution, most previous TTA methods assume that the model is trained on class-balanced data.
This assumption overlooks another type of distribution shift, known as \emph{label shift}, in which label distribution varies between source $p_s(y)$ and target $p_t(y)$.
Label shift has been studied extensively in the long-tailed recognition literature~\cite{kang2019decoupling,cao2019learning,cui2019class,hong2021disentangling}.
Considering only one type of distribution shifts in real-world scenarios is infeasible, as both covariate and label shifts frequently occur simultaneously~\cite{liu2021adversarial,li2021imbalanced}.
For example, different object classes such as buildings and trees are more prevalent in certain environments, such as cities or forests.

We found that the majority of existing TTA approaches fail to adapt the pre-trained model when it is trained on a long-tailed dataset.
This is because most TTA methods that employ entropy minimization~\cite{wang2020tent,niu2022efficient,gong2022note,choi2022improving,lim2023ttn} are flawed due to the model bias towards the dominant classes in source data.
In other words, since the predictions on test samples are often biased toward the majority classes of training data, utilizing entropy minimization would lead the model to increase its confidence in predicting the dominant classes.
While some recent TTA studies have addressed similar challenging issues, such as temporally correlated test data~\cite{gong2022note} and class-imbalanced test samples~\cite{zhao2023delta}, they do not cover the situation where the source domain data has a long-tailed label distribution, which can lead to bias in the source model. 
Furthermore, we observed that training the source model on a long-tailed dataset with long-tailed recognition techniques, such as balanced softmax~\cite{ren2020balanced}, is insufficient to stabilize TTA algorithms, as shown in Fig.~\ref{fig:motivation}.

To tackle such a non-trivial issue, we propose a novel label shift adapter, which is designed to adapt the pre-trained model according to the label distribution during inference.
Before deploying the model to the server, we train the label shift adapter with a pre-trained source model, which takes the label distribution as input.
The label shift adapter is optimized to produce the optimal parameters according to the label distribution.
To make our method applicable to any model architecture, we design the label shift adapter to predict only the parameters associated with a part of the source model.
After model deployment, we estimate the label distribution of target domain data for injecting the appropriate input into the label shift adapter during inference.
Moreover, our proposed method can be easily integrated with TTA methods such as TENT~\cite{wang2020tent} and IABN~\cite{gong2022note} to adapt the model to the target domain.
Combining TTA approaches with the proposed label shift adapter enables robust model adaptation to the target domain, even in the presence of covariate and label shifts simultaneously.
Through extensive experiments, we demonstrate that our method outperforms the existing TTA methods when both source and target domain datasets have class-imbalanced label distributions.

In summary, the main contributions are as follows:
\begin{itemize}
    \setlength{\itemsep}{-2.5pt}
    \item We introduce a novel label shift adapter that produces the optimal parameters according to the label distribution. By utilizing the label shift adapter, we can develop a robust TTA algorithm that can handle both covariate and label shifts simultaneously.
    \item Our approach is easily applicable to any model regardless of the model architecture and pre-training process. It can be simply integrated with other TTA algorithms.
    \item Through extensive experiments on six benchmarks, we demonstrate that our method enhances the performance significantly when source and target domain datasets have class-imbalanced label distributions.
\end{itemize}

\section{Related Work}

\noindent\textbf{Source-Free Domain Adaptation.}
Unsupervised domain adaptation (UDA) methods have been widely applied in cross-domain applications such as classification~\cite{ganin2015unsupervised,ganin2016domain}, object detection~\cite{yoo2022unsupervised}, and semantic segmentation~\cite{zou2018unsupervised}.
However, UDA approaches require access to both source and target domains simultaneously.
This restriction makes these approaches frequently impractical due to computational costs and data privacy concerns.
On the other hand, source-free domain adaptation (DA)~\cite{liang2020we,kurmi2021domain} overcomes this limitation by adapting a pre-trained model to the target domain using only unlabeled target data.
However, existing source-free DA methods barely consider label shifts, which limits their applicability in real-world scenarios.

\noindent\textbf{Domain Adaptation for Label Shift.}
Several methods~\cite{azizzadenesheli2019regularized,lipton2018detecting,tan2020class, jiang2020implicit,liu2021adversarial,li2021imbalanced} have been developed to investigate a more realistic scenario of domain adaptation in which covariate and label shifts co-occur.
To alleviate label shift, previous studies employ the re-weighting method~\cite{azizzadenesheli2019regularized,lipton2018detecting} by estimating target domain label distribution.
Recent approaches employ an alternative training scheme~\cite{liu2021adversarial} and a secondary pseudo label~\cite{li2021imbalanced} to alleviate label shifts in unsupervised domain adaptation.
However, it is challenging to adapt the model to the target domain in real time using these methods.
Therefore, we focus on addressing such co-existence of covariate and label shifts in the TTA setting.

\noindent\textbf{Test-Time Adaptation.}
Fully test-time adaptation~\cite{wang2020tent} aims to improve model performance on target domain data through adaptation with unlabeled test samples during inference.
Previous work~\cite{schneider2020improving} improves the robustness under covariate shift by using the statistics of test batch in normalization layer.
TENT~\cite{wang2020tent} further optimizes affine parameters of batch normalization layers using entropy minimization.
Before the pre-trained model deployment to the server, several approaches train additional modules to appropriately interpolate training and test statistics~\cite{zou2022learning,lim2023ttn} or regularize the model parameters~\cite{choi2022improving} for TTA.
Recent several studies~\cite{boudiaf2022parameter,gong2022note,zhao2023delta} address the model to be more robust under non-i.i.d or class-imbalanced test samples.
However, they assume that the source domain datasets are balanced, where the pre-trained model is not biased towards the dominant classes due to the class-imbalanced label distribution.
Different from the existing studies, our research tackles the cases in which both source and target domain datasets are class-imbalanced, which is more challenging and practical.

\begin{figure*}[t!]
    \centering
    \includegraphics[width=0.9\linewidth]{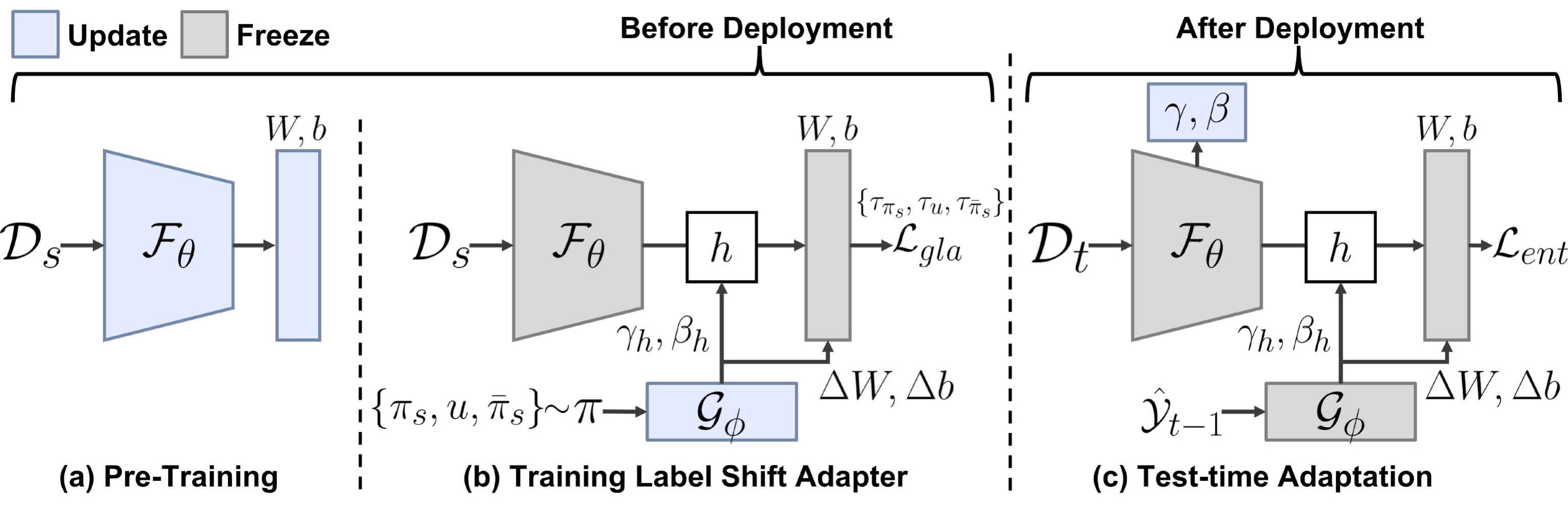}
    \vspace{-0.2cm}
    \caption{\textbf{Overview of the proposed method.} (a) We take a frozen pre-trained source model trained with $\mathcal{D}_{s}$. (b) Before model deployment, we train a label shift adapter with a frozen source pre-trained model. (c) After model deployment, we adapt the model to the target domain by integrating the label shift adapter into other TTA algorithms.}
    \vspace{-0.5cm}
    \label{fig:model}
\end{figure*}

\noindent\textbf{Long-Tailed Recognition.}
It is natural that datasets have long-tailed distributions in the real world.
Previous studies addressed this issue by altering loss functions~\cite{lin2017focal,cui2019class,cao2019learning,samuel2021distributional,lee2021improving}, adjusting logits~\cite{menon2020long,ren2020balanced}, and employing multiple experts that are specialized in different label distributions~\cite{xiang2020learning,zhou2020bbn,zhang2022self}.
Recently, several studies such as LADE~\cite{hong2021disentangling}, SADE~\cite{zhang2022self}, and BalPoE~\cite{aimar2022balanced} have introduced test-agnostic long-tailed recognition, where the training label distribution is long-tailed while the test label distribution is agnostic.
However, LADE needs the true test label distribution to adjust the logits, and SADE and BalPoE require multiple expert architectures.
Due to these aspects, they are difficult to apply the TTA algorithms.
Inspired by such studies, we design a novel label shift adapter, which has the capability to handle unknown test label distribution using training long-tailed distribution.
In contrast to previous methods, our method is applicable to any model regardless of its architecture and can be employed without true test label distribution.
In this paper, we focus on handling the label shifts in TTA setting.

\noindent\textbf{Predicting Weights.}
A hypernetwork~\cite{ha2016hypernetworks} is a deep neural network to produce the weights of another neural network.
Hypernetworks have been developed for federated learning~\cite{shamsian2021personalized}, multi-task learning~\cite{navon2021learning,lin2020controllable,mahabadi2021parameter}, and continuous learning~\cite{von2020continual,brahma2021hypernetworks}.
Moreover, adapter layers between existing layers of the model have been proposed for fine-tuning~\cite{lin2020exploring,hu2021lora}.
The key functional difference is that our method produces the parameters to handle the label shifts.

\section{Method}

\subsection{Problem Formulation}

In the TTA task, labeled samples from the source domain $\mathcal{D}_{s}=\{ (x,y) \sim p_s(x,y) \}$ and unlabeled samples from the target domain $\mathcal{D}_{t}=\{ x \sim p_t(x) \}$.
TTA aims to predict the labels of target domain samples by updating the source model to the target model during inference.
Specifically, under the TTA scheme, the model receives a mini-batch $x_t$ of test samples in the $t$-th inference step.

Generally, previous TTA literature assumes only covariate shift, where $p_s(x) \neq p_t(y)$.
In other words, the existing TTA methods only consider class-balanced datasets in training and testing.
Different from previous works, we assume the joint presence of covariate and label shifts~\cite{jiang2020implicit,tan2020class,li2021imbalanced}: $p_s(x) \neq p_t(x)$ and $p_s(y) \neq p_t(y)$, which is more practical and natural in real-world scenarios.
In particular, when $p_s(y)$ has long-tailed label distribution, TTA methods are flawed, despite leveraging long-tailed recognition methods.
It is due to the fact that most TTA methods are not able to reduce the model's bias toward the majority classes.
Our goal is to design a novel method for TTA that can perform stably regardless of $p_s(y)$ and $p_t(y)$, while the model can be adapted during inference.

\subsection{Label Shift Adapter}

In this paper, we propose a novel label shift adapter for TTA to handle label distribution shifts in TTA.
Entropy minimization~\cite{wang2020tent,niu2022efficient,choi2022improving,gong2022note,lim2023ttn} is widely used for TTA to optimize the model with unlabeled test samples during inference.
Intuitively, entropy minimization makes individual predictions confident.
During test time, entropy minimization loss is utilized as follows:
\begin{equation}
    \mathcal{L}_{ent} = - \sum_{x \sim p_t(x)} f(x) \log f(x),
\end{equation}
where $f$ denotes a model $f: x \rightarrow y$.
However, if the pre-trained source model $f(x)$ is biased toward the majority classes due to the long-tailed label distribution of $\mathcal{D}_{s}$, the predictions on test samples also would be biased towards the majority classes in $\mathcal{D}_{s}$ regardless of label distribution of $\mathcal{D}_{t}$. 
In other words, it is not appropriate to minimize under the shifted label distribution because the model prediction estimates $p_s(y|x)$, which is strongly coupled with $p_s(y)$ and may differ from $p_t(y)$, as explained by the Bayes' rule~\cite{hong2021disentangling}:
\begin{equation}
    p_s(y|x) = \frac{p_s(y)p_s(x|y)}{p_s(x)} = \frac{p_s(y)p_s(x|y)}{\sum_c p_s(c)p_s(x|c)},
\end{equation}
where $c$ denotes the class index.

To address diverse $p_t(y)$ trained with a long-tailed distribution $p_s(y)$, recent long-tailed recognition methods~\cite{zhang2022self,aimar2022balanced} have proposed the training strategy utilizing the multiple diverse experts, which are specialized in handling different label distributions, such as long-tailed and uniform label distributions.
However, these approaches are not directly applicable to TTA setting, as they are designed for multiple-expert model architectures.
Inspired by this strategy, we aim to develop the model $f$ that is suitable for TTA by dynamically adapting the model $f$ to diverse target label distributions $p_t(y)$.
Therefore, we introduce a novel label shift adapter that can produce the optimal parameters according to label distribution during inference while it is applicable to any model regardless of model architecture.

Fig.~\ref{fig:model} shows the overview of the proposed method.
Our method consists of three stages.
First, our method takes the pre-trained source model in an off-the-shelf manner.
Before model deployment, we train the label shift adapter with the frozen pre-trained model, which produces optimal parameters depending on the label distribution.
Then, our label shift adapter can be integrated into other TTA algorithms, such as TENT~\cite{wang2020tent}, after model deployment.
In specific, we optimize the affine parameters in the normalization layers of a feature extractor while adapting the model to the target label distribution $p_t(y)$ by estimating the label distribution of $\mathcal{D}_t$ during inference.

\noindent\textbf{Label Shift Adapter.}
Before training a label shift adapter, we pre-train a model $f: x \rightarrow y$ using a source domain data $\mathcal{D}_{s}$, where the model consists of a feature extractor $\mathcal{F}_{\theta}$ and a classifier weights $W \in \mathbb{R}^{d \times C}, b \in \mathbb{R}^{1 \times C}$.
$C$ and $d$ denote the number of classes and channel of the output $h \in \mathbb{R}^{1 \times d}$ of the feature extractor, respectively.
As several TTA methods~\cite{choi2022improving,zou2022learning,lim2023ttn} include an additional stage for training extra components, the label shift adapter is trained with the frozen pre-trained model before model deployment.

The label shift adapter $\mathcal{G}_{\phi}$ receives the label distribution $\pi \in \mathbb{R}^{C}$ as conditional input.
For applicability and efficiency, we design the label shift adapter to generate the parameters for a part of the model.
With $\pi$, the label shift adapter $\mathcal{G}_{\phi}$ predicts affine parameters $\gamma_h \in \mathbb{R}^{1 \times d}, \beta_h \in \mathbb{R}^{1 \times d}$ and the weight difference $\Delta W \in \mathbb{R}^{d \times C}, \Delta b \in \mathbb{R}^{1 \times C}$ for the classifier weights $W, b$.
The affine parameters $\gamma_h$ and $\beta_h$ are applied to the hidden feature map $h$, which is the output of the feature extractor: $h=\mathcal{F}_\theta(x)$.
Then, we compute the output $\hat{y}$ using $W + \Delta W$ and $b + \Delta b$.
Formally, $\hat{y}$ is computed in the classifier layer as follows:
\begin{equation}\label{eq:adapter forward}
    \hat{y} = (\gamma_h h + \beta_h) \cdot (W + \Delta W) + (b + \Delta b).
\end{equation}

\begin{figure}
    \centering
    \includegraphics[width=0.8\linewidth]{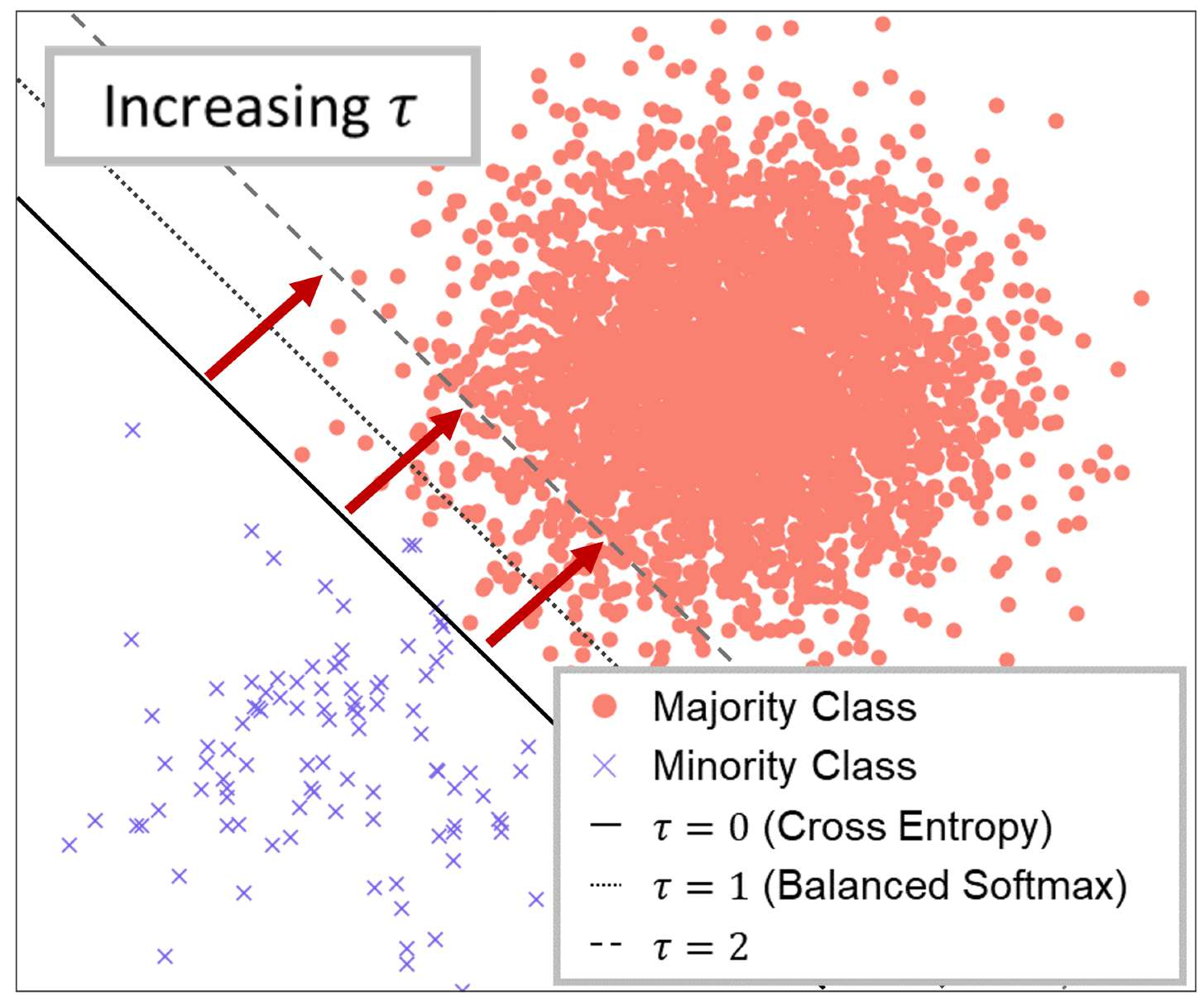}
    \caption{Visualization for understanding $\tau$ in a generalized logit adjusted loss. By adjusting $\tau$, we can control the bias during training. As $\tau$ increases, the decision boundary }
    \vspace{-0.3cm}
    \label{fig:tau}
\end{figure}

\noindent\textbf{Objective Function.}
The label shift adapter aims to create the optimal parameters depending on the label distribution $\pi$.
During training the label shift adapter, we sample $\pi$ based on $\pi_s$, which is the label distribution of $\mathcal{D}_{s}$.

To optimize the label shift adapter, we employ a generalized logit adjusted loss~\cite{menon2020long,aimar2022balanced}, which incorporates a controlled bias during training:
\begin{equation}
    \mathcal{L}_{gla} = - \sum_{(x_i, y_i) \sim \mathcal{D}_s} y_i \log \sigma ( \hat{y}_i + \tau \log \pi_s ),
\end{equation}
where $\sigma$ denotes a softmax function.
$\hat{y}_i$ indicates the output logits before computing the softmax function.
$\tau \in \mathbb{R}^{1}$ is a scalar value for controlling the bias towards different parts of the label distribution.
We sample $\pi$ at each iteration during the label shift adapter training.
Specifically, the label distribution $\pi$ is sampled from three types of label distribution $\{ \pi_s, u, \bar{\pi}_s \}$, where $\bar{\pi}_s$ indicate inverse label distribution that is obtained by inverting the order of training label distribution $\pi_s$.
$u$ denotes uniform label distribution.
We select the appropriate $\tau$ for sampled label distribution $\pi$, where we set the hyperparameter $\tau$ matching each $\pi \subset \{ \pi_s, u, \bar{\pi}_s \}$.
For example, if $\pi_s$ is sampled for $\pi$, $\tau$ is set to 0, resulting in the use of a cross-entropy loss. 
On the other hand, $\tau$ is set to 1 and 2 for $u$ and $\bar{\pi}_s$, respectively, which correspond to the balanced softmax~\cite{ren2020balanced} and inverse softmax~\cite{zhang2022self} that simulate uniform and inverse label distributions.
Intuitively, as $\tau$ increases, the decision boundary moves away from minority classes and towards the majority classes~\cite{menon2020long,aimar2022balanced}, as shown in Fig.~\ref{fig:tau}.
Technically, it is possible to sample $\pi$ in continuous label space instead of discrete label distributions.
However, we found that sampling three distinctive label distributions is empirically sufficient to train the label shift adapter.

\noindent\textbf{Test-Time Adaptation.}
After model deployment, we adapt the pre-trained model $f$ using test samples $x_t$ at $t$-th inference step.
In specific, our method updates affine parameters $\gamma, \beta$ in normalization layers of feature extractor $F_{\theta}$.
With $t$-th test samples $x_t$, the model minimizes the prediction entropy, following the previous work~\cite{wang2020tent}.

However, minimizing the entropy of the predictions $\hat{y}_t$ is insufficient to adapt the model when label shift occurs.
Therefore, the label shift adapter creates a part of parameters (\textit{i.e.}, $\gamma_h, \beta_h, \Delta W$, and $\Delta b$) of $f$ depending on estimated label distribution $\hat{\mathcal{Y}}$.
To estimate the label distribution $\hat{\mathcal{Y}}$ of $D_t$, we employ an exponential moving average.
To be specific, the estimated target label distribution $\hat{\mathcal{Y}}_t$ at $t$-th step is updated recursively:
\begin{equation}
    \hat{\mathcal{Y}}_t = 
        \begin{cases}
            u, & \text{if $t=0$} \\ 
            \alpha \bar{y}_t + (1 - \alpha) \hat{\mathcal{Y}}_{t-1}, & \text{if $t>0$}
        \end{cases},
\end{equation}
where $\alpha \in [ 0, 1 ]$ denotes the momentum hyper-parameter, $\bar{y}_t = \frac{1}{n_t} \sum_{i=1}^{n_t} \hat{y}^i_t$ is the average model prediction on test samples $x_t$ at the $t$-th step.
We initialize $\hat{\mathcal{Y}}$ as $u \in \mathbb{R}^{C}$, which is a uniform distribution vector: $u[c] = \frac{1}{C}, c=1,2, \cdots, C$.
Based on estimated target label distribution $\hat{\mathcal{Y}}_{t-1}$, the label shift adapter produces the optimal parameters.
With $\hat{\mathcal{Y}}_{t-1}$, we can adapt the model to the target domain using the refined entropy minimization as follows:
\begin{equation}
    \mathcal{L}_{ent} = - \sum_{x \sim p_t(x)} f(x;\mathcal{G}_{\phi}(\hat{\mathcal{Y}})) \log f(x;\mathcal{G}_{\phi}(\hat{\mathcal{Y}})),
\end{equation}
where the estimated label distribution $\hat{\mathcal{Y}}$ is fed into the label shift adapter $\mathcal{G}_{\phi}$ to handle label distribution shift, and the generated parameters from $\mathcal{G}_{\phi}$ adjusts the model $f$ as shown in Eq.~\ref{eq:adapter forward}.
Intuitively, while our strategy adapts the model to the target label distribution, entropy minimization boosts the confidence of correct classes.

\begin{table*}[t!]
    \scriptsize
    \centering
    \begin{tabular}{@{}l@{ }ccccccc@{\quad}c@{\;}|ccccccc@{\quad}c@{}}
        \toprule
        \multirowcell{3}{\textbf{Method}} & \multicolumn{8}{c|}{\textbf{CIFAR-10-C}} & \multicolumn{8}{c}{\textbf{CIFAR-100-C}} \\
        & \multicolumn{3}{c}{\textbf{Forward-LT}} & \textbf{Uni.} & \multicolumn{3}{c}{\textbf{Backward-LT}} & \multirowcell{2}{\multicolumn{1}{c}{\textbf{Avg.}}} &
        \multicolumn{3}{c}{\textbf{Forward-LT}} & \textbf{Uni.} & \multicolumn{3}{c}{\textbf{Backward-LT}} & \multirowcell{2}{\multicolumn{1}{c}{\textbf{Avg.}}} \\
        \cmidrule(lr){2-4} \cmidrule(lr){5-5} \cmidrule(lr){6-8}
        \cmidrule(lr){10-12} \cmidrule(lr){13-13} \cmidrule(lr){14-16}
        & 50 & 25 & 10 & 1 & 10 & 25 & 50 && 50 & 25 & 10 & 1 & 10 & 25 & 50 \\
        \midrule
        Source & 56.78 & 55.27 & 52.82 & 44.74 & 37.28 & 34.78 & 33.88 & 45.08
                & 33.53 & 31.95 & 29.36 & 22.14 & 14.98 & 12.37 & 11.12 & 22.21 \\
        \midrule
        BN Stats & 78.60 & 76.21 & 71.77 & 53.19 & 35.00 & 28.62 & 25.18 & 52.65 
                  & 49.03 & 46.61 & 42.92 & 31.30 & 20.07 & 16.09 & 13.86 & 31.41 \\
        ONDA & 77.93 & 75.94 & 72.34 & 55.28 & 37.71 & 31.76 & 28.64 & 54.23 
              & 48.30 & 46.52 & 42.82 & 31.77 & 20.58 & 16.55 & 14.41 & 31.57 \\
        PseudoLabel & 79.39 & 76.40 & 71.37 & 47.90 & 30.68 & 24.95 & 22.61 & 50.47 
                      & 50.80 & 48.24 & 43.56 & 25.51 & 17.03 & 14.28 & 11.99 & 30.20 \\
        LAME & 58.27 & 55.88 & 52.27 & 41.60 & 34.14 & 32.15 & 31.54 & 43.69 
              & 32.63 & 30.99 & 28.12 & 20.81 & 13.94 & 11.58 & 10.29 & 21.19 \\
        CoTTA & 80.45 & 77.86 & 73.12 & 53.75 & 33.89 & 27.80 & 24.01 & 52.98 
               & 48.32 & 45.71 & 42.30 & 30.42 & 21.65 & 18.14 & 16.33 & 31.84 \\
        NOTE & 76.27 & 74.79 & 72.18 & 61.98 & 53.13 & 50.94 & 49.45 & 62.68
              & 44.52 & 43.15 & 40.52 & 34.25 & 23.92 & 20.60 & 19.11 & 32.30 \\
        \midrule
        TENT & 80.36 & 76.99 & 71.28 & 43.65 & 28.64 & 23.01 & 20.32 & 49.18 
              & 51.74 & 49.24 & 44.05 & 21.30 & 15.86 & 13.40 & 11.25 & 29.55 \\
        \rowcolor{gray!10}
        +Ours & 80.39 & 78.03 & 73.35 & 53.91 & 37.85 & 32.83 & 30.32 & 55.24
                & \textbf{52.43} & \textbf{50.17} & \textbf{46.07} & 33.27 & 21.23 & 17.12 & 15.13 & 33.63 \\
        \rowcolor{gray!10} 
            & \textcolor{mblue}{+0.03} & \textcolor{mblue}{+1.04} & \textcolor{mblue}{+2.06} & \textcolor{mblue}{+10.25} & \textcolor{mblue}{+9.21} & \textcolor{mblue}{+9.82} & \textcolor{mblue}{+10.00} & \textcolor{mblue}{+6.06}
            & \textcolor{mblue}{+0.69} & \textcolor{mblue}{+0.93} & \textcolor{mblue}{+2.02} & \textcolor{mblue}{+11.98} & \textcolor{mblue}{+5.37} & \textcolor{mblue}{+3.72} & \textcolor{mblue}{+3.87} & \textcolor{mblue}{+4.08} \\
        \midrule
        IABN & 76.23 & 74.84 & 72.22 & 62.22 & 53.29 & 50.88 & 49.69 & 62.77
               & 44.79 & 43.24 & 40.63 & 34.01 & 23.95 & 20.67 & 19.16 & 32.35 \\
        \rowcolor{gray!10} 
        +Ours & \textbf{80.58} & \textbf{78.62} & \textbf{75.26} & \textbf{63.34} & \textbf{68.54} & \textbf{70.07} & \textbf{71.64} & \textbf{72.58}
                & 52.06 & 49.71 & 46.03 & \textbf{36.84} & \textbf{29.29} & \textbf{26.33} & \textbf{25.50} & \textbf{37.97} \\
        \rowcolor{gray!10} 
            & \textcolor{mblue}{+4.35} & \textcolor{mblue}{+3.78} & \textcolor{mblue}{+3.04} & \textcolor{mblue}{+1.12} & \textcolor{mblue}{+15.24} & \textcolor{mblue}{+19.20} & \textcolor{mblue}{+21.95} & \textcolor{mblue}{+9.81}
            & \textcolor{mblue}{+7.26} & \textcolor{mblue}{+6.47} & \textcolor{mblue}{+5.40} & \textcolor{mblue}{+2.83} & \textcolor{mblue}{+5.34} & \textcolor{mblue}{+5.67} & \textcolor{mblue}{+6.33} & \textcolor{mblue}{+5.62} \\
        \bottomrule
    \end{tabular}
    \caption{\textbf{Comparison of accuracy on CIFAR-10-C and CIFAR-100-C.} The source model is trained with CIFAR-10-LT and CIFAR-100-LT. We report the average accuracy of 15 corruption types on various test label distributions. Uni. indicates the uniform distribution. Numbers under Forward-LT and Backward-LT denote the imbalance ratio. We integrate our method into TENT~\cite{wang2020tent} and IABN~\cite{gong2022note}.}
    \label{Table:experiment-cifar}
\end{table*}

\begin{table*}[t!]
    \small
    \centering
    \begin{tabular}{lcccccccc}
        \toprule
        \multirowcell{2}{\textbf{Method}} & \multicolumn{3}{c}{\textbf{Forward-LT}} & \textbf{Uni.} & \multicolumn{3}{c}{\textbf{Backward-LT}} & \multirowcell{2}{\multicolumn{1}{c}{\textbf{Avg.}}} \\
        \cmidrule(lr){2-4} \cmidrule(lr){5-5} \cmidrule(lr){6-8}
        & 50 & 25 & 10 & 1 & 10 & 25 & 50 \\
        \midrule
        Source & 26.15 & 25.64 & 24.67 & 21.46 & 18.28 & 17.07 & 16.56 & 21.40 \\
        \midrule
        BN Stats & 39.47 & 38.89 & 37.71 & 33.63 & 29.48 & 28.07 & 27.20 & 33.49 \\
        ONDA & 39.45 & 38.83 & 37.71 & 33.56 & 29.33 & 28.01 & 26.96 & 33.41 \\
        PseudoLabel & 41.46 & 40.78 & 39.31 & 33.49 & 29.79 & 28.36 & 27.67 & 34.41 \\
        LAME & 26.08 & 25.57 & 24.58 & 21.37 & 18.20 & 17.01 & 16.48 & 21.33 \\
        CoTTA & 40.22 & 39.81 & 39.10 & 35.40 & 30.21 & 28.72 & 27.65 & 34.44 \\
        NOTE & 42.43 & 41.65 & 40.36 & 35.17 & 30.99 & 29.14 & 28.17 & 35.41 \\
        \midrule
        TENT & 39.40 & 38.73 & 37.27 & 29.05 & 29.31 & 28.26 & 27.28 & 32.76 \\
        \rowcolor{gray!10} +Ours & 44.52 & 43.03 & 40.86 & 34.18 & 31.32 & 31.21 & 31.28 & 36.63 \\
        \rowcolor{gray!10} & \textcolor{mblue}{+5.12} & \textcolor{mblue}{+4.30} & \textcolor{mblue}{+3.59} & \textcolor{mblue}{+5.14} & \textcolor{mblue}{+2.01} & \textcolor{mblue}{+2.94} & \textcolor{mblue}{+3.99} & \textcolor{mblue}{+3.87} \\
        
        \midrule
        IABN & 42.44 & 41.69 & 40.39 & 35.20 & 31.02 & 29.20 & 28.22 & 35.45 \\
        \rowcolor{gray!10} +Ours & \textbf{46.88} & \textbf{45.16} & \textbf{42.68} & \textbf{35.72} & \textbf{33.18} & \textbf{32.91} & \textbf{33.17} & \textbf{38.53} \\
        \rowcolor{gray!10} & \textcolor{mblue}{+4.44} & \textcolor{mblue}{+3.47} & \textcolor{mblue}{+2.29} & \textcolor{mblue}{+0.52} & \textcolor{mblue}{+2.16} & \textcolor{mblue}{+3.71} & \textcolor{mblue}{+4.95} & \textcolor{mblue}{+3.08} \\
        \bottomrule
    \end{tabular}
    \caption{\textbf{Comparison of accuracy on ImageNet-C.} We report the average accuracy of 15 corruption types on various test label distributions. Uni. indicates the uniform distribution. Numbers under Forward-LT and Backward-LT denote the imbalance ratio.}
    \vspace{-0.3cm}
    \label{Table:experiment-imagenet}
\end{table*}




\noindent\textbf{Techniques for Label Shift Adapter.}
In long-tailed recognition literature~\cite{kang2019decoupling,alshammari2022long}, it is known that the classifier layer plays an essential role in resolving the label shifts. 
Based on this intuition, we have designed the label shift adapter to produce the parameters only for the parts associated with the classifier layer.
Specifically, $\Delta W$ and $\Delta b$ is the weight difference of $W$ and $b$, and $\gamma_h$ and $\beta_h$ shift the feature vector $h$ properly.
By predicting only a small portion of the model instead of predicting the entire model weights, our label shift adapter offers various benefits.
Our label shift adapter is readily applicable to any pre-trained models, regardless of the model architecture.
Moreover, this strategy is computationally efficient, as the label shift adapter requires negligible extra computational costs.

We discovered that it is more effective to utilize a mapping vector $m \in \mathbb{R}^{C}$ to make the label distribution $\pi$ a scalar, instead of directly using label distribution $\pi$ as the input of the label shift adapter.
In specific, $m^\intercal \pi \in \mathbb{R}^{1}$ is fed into label shift adapter, instead of $\pi$.
Here, the mapping vector is the class-wise coefficients that increase in proportion to the order of class frequency in the training set.
Empirically, this strategy makes the label shift adapter training more stable.
Instead of using a complex label space as a condition, this technique feeds the degree of imbalance to label shift adapter, which can easily interpret the condition.

\noindent\textbf{Rationale behind Label Shift Adapter.}
Label shift adapter plays a crucial role in handling label shifts by adjusting the parameters based on the estimated label distribution at test time.
Since the label distribution in the target domain is unknown, it is possible to train the label shift adapter by simulating various label distributions using the source domain dataset.
By virtue of this process, the label shift adapter can learn to produce appropriate parameters based on the label distribution.
Since the label distribution in the target domain is unknown, it is possible to train the adapter by simulating various label distributions using the source domain dataset.
Moreover, this approach also allows for the effective handling of biases caused by the label distribution of the source domain dataset.
This is why transferring the model from the source domain to the target domain proves to be an effective approach.

\section{Experiments}

\subsection{Experiment Setup}

We evaluate the effectiveness of our proposed method on six datasets widely used in domain adaptation literature.
In contrast to the traditional TTA setting, we utilize imbalanced versions of the datasets for training and evaluation.

\noindent\textbf{CIFAR-10\&100 and ImageNet~\cite{cao2019learning,hendrycks2018benchmarking}.}
For the training set of the source domain, we utilize CIFAR-10-LT, CIFAR-100-LT~\cite{cao2019learning}, and ImageNet-LT~\cite{liu2019large}, which are the long-tailed version of CIFAR-10, CIFAR-100 and ImageNet, following the protocol in long-tailed recognition work.
Note that the imbalance ratio $\rho$ is defined as $\rho = \frac{\max_i n_i}{\min_i n_i}$, where $n_i$ denotes the number of class $i$ samples in the dataset.
In our experiments, the imbalance ratio of CIFAR-10-LT and CIFAR-100-LT is set to 100.
ImageNet-LT is obtained by sampling from ImageNet using a Pareto distribution with $\alpha=6$~\cite{liu2019large}, following the previous work.
The categories in the training set of ImageNet-LT contain between 5 and 1280 samples, with an imbalanced ratio set to 256.
For evaluation, we utilize three corrupted test sets: CIFAR-10-C, CIFAR-100-C, and ImageNet-C~\cite{hendrycks2018benchmarking}, which consist of 15 corruption types at five severity levels.
The severity level is set to 5 and 3 for CIFAR-C and ImageNet-C, respectively.

Following the test-agnostic long-tailed recognition setting~\cite{hong2021disentangling,zhang2022self}, the models are evaluated on multiple subsets of test datasets that follow different label distributions.
We construct three types of test label distributions as follows: (i) Forward distribution: as the imbalance ratio increases, it becomes similar to the training label distribution. (ii) Uniform distribution: a class-balanced test dataset. (iii) Backward distribution: the order of classes is reversed, causing it to deviate more from the training distribution, as the imbalance ratio increases.
Note that the degree of label shifts increases from Forward to Backward.

\begin{table*}[t!]
    \parbox{0.3\textwidth}{
        \small
        \centering
        \begin{tabular}{lp{0.1\textwidth}<{\centering}}
            \toprule
            \multicolumn{1}{c}{\textbf{Method}} & \textbf{VISDA-C} \\
            \midrule
            Source & 51.45 \\
            \midrule
            BN Stats & 49.33 \\
            ONDA & 50.68 \\
            PseudoLabel & 49.50 \\
            LAME & 50.72 \\
            CoTTA & 49.88 \\
            NOTE & 49.37 \\
            \midrule
            TENT & 48.68 \\
            \rowcolor{gray!10} +Ours & \textbf{72.97} \\
            \rowcolor{gray!10} & \textcolor{mblue}{+24.29} \\
            \bottomrule
        \end{tabular}
        \caption{\textbf{Comparison of accuracy on VisDA-C (RSUT)}.}
        \label{Table:experiment-visda}
    }
    \hfill
    \parbox{0.68\textwidth}{
        \small
        \centering
        \begin{tabular}{lp{0.05\textwidth}<{\centering}p{0.05\textwidth}<{\centering}p{0.05\textwidth}<{\centering}p{0.05\textwidth}<{\centering}p{0.05\textwidth}<{\centering}p{0.05\textwidth}<{\centering}p{0.05\textwidth}<{\centering}}
            \toprule
            \multicolumn{1}{c}{\textbf{Method}} & \textbf{C$\rightarrow$P} & \textbf{C$\rightarrow$R} & \textbf{P$\rightarrow$C} & \textbf{P$\rightarrow$R} & \textbf{R$\rightarrow$C} & \textbf{R$\rightarrow$P} & \textbf{Avg.} \\
            \midrule
            Source & 45.39 & 44.53 & 32.94 & 64.33 & 40.22 & \textbf{68.92} & 49.39 \\
            \midrule
            BN Stats & 44.30 & 48.27 & 35.63 & 62.17 & 40.73 & 62.20 & 48.88 \\
            ONDA & 44.84 & 47.57 & 35.20 & 62.09 & 40.61 & 63.83 & 49.02 \\
            PseudoLabel & 47.98 & 49.34 & 37.71 & 62.42 & 39.38 & 63.21 & 50.01 \\
            LAME & 41.68 & 42.27 & 32.40 & 63.57 & 37.92 & 66.94 & 47.46 \\
            CoTTA & 44.46 & 48.19 & 35.63 & 62.34 & 40.73 & 62.20 & 48.92 \\
            NOTE & 43.02 & 42.38 & 38.64 & 61.69 & 41.40 & 64.33 & 48.58 \\
            \midrule
            TENT & \textbf{49.60} & 49.51 & \textbf{38.96} & 63.08 & 41.25 & 64.52 & 51.15 \\
            \rowcolor{gray!10} +Ours & \textbf{49.60} & \textbf{53.13} & 37.81 & \textbf{66.45} & \textbf{41.35} & 68.35 & \textbf{52.78} \\
            \rowcolor{gray!10} & 0.00 & \textcolor{mblue}{+3.62} & -1.15 & \textcolor{mblue}{+3.37} & \textcolor{mblue}{+0.10} & \textcolor{mblue}{+3.83} & \textcolor{mblue}{+1.63} \\
            \bottomrule
        \end{tabular}
        \caption{\textbf{Comparison of accuracy on three domains of Officehome (RSUT)}: \textbf{C}: Clipart, \textbf{P}: Product, \textbf{R}: Realworld. }
        \label{Table:experiment-officehome}
    }
    \vspace{-0.2cm}
\end{table*}








\begin{table*}[t!]
    \small
    \centering
    \begin{tabular}{lp{0.04\textwidth}<{\centering}p{0.04\textwidth}<{\centering}p{0.04\textwidth}<{\centering}p{0.04\textwidth}<{\centering}p{0.04\textwidth}<{\centering}p{0.04\textwidth}<{\centering}p{0.04\textwidth}<{\centering}p{0.04\textwidth}<{\centering}p{0.04\textwidth}<{\centering}p{0.04\textwidth}<{\centering}p{0.04\textwidth}<{\centering}p{0.04\textwidth}<{\centering}p{0.04\textwidth}<{\centering}}
        \toprule
        \multicolumn{1}{c}{\textbf{Method}} & \textbf{C$\rightarrow$P} & \textbf{C$\rightarrow$R} & \textbf{C$\rightarrow$S} & \textbf{P$\rightarrow$C} & \textbf{P$\rightarrow$R} & \textbf{P$\rightarrow$S} & \textbf{R$\rightarrow$C} & \textbf{R$\rightarrow$P} & \textbf{R$\rightarrow$S} & \textbf{S$\rightarrow$C} & \textbf{S$\rightarrow$P} & \textbf{S$\rightarrow$R} & \textbf{Avg.} \\
        \midrule
        Source & 52.73 & 74.87 & 52.15 & 58.42 & 81.22 & 61.82 & 66.03 & 69.58 & 55.31 & 63.92 & 59.68 & 75.43 & 64.26 \\
        \midrule
        BN Stats & 56.81 & 77.05 & 54.10 & 63.63 & 81.12 & 60.22 & 67.38 & 70.00 & 56.84 & 71.75 & 68.72 & 80.18 & 67.32 \\
        ONDA & 56.82 & 78.32 & 54.81 & 63.99 & 81.79 & 61.86 & 67.14 & 70.09 & 58.11 & 71.60 & 69.34 & 80.77 & 67.89 \\
        PseudoLabel & 61.81 & 77.43 & 56.25 & 62.56 & 81.64 & 62.04 & 71.06 & 73.89 & 58.49 & 71.81 & 70.38 & 80.12 & 68.96 \\
        LAME & 49.20 & 72.45 & 48.69 & 57.81 & 80.09 & 60.85 & 65.25 & 68.19 & 53.97 & 61.00 & 55.66 & 73.25 & 62.20 \\
        CoTTA & 56.88 & 77.33 & 54.18 & 63.69 & 81.31 & 60.26 & 67.44 & 70.07 & 57.14 & 71.69 & 68.85 & 80.56 & 67.45 \\
        NOTE & 55.38 & 74.15 & 57.98 & 65.59 & 81.66 & 64.65 & 71.29 & 73.32 & 63.28 & 72.28 & 68.31 & 80.25 & 69.01 \\
        \midrule
        TENT & \textbf{63.26} & 77.10 & 59.76 & 66.69 & 80.02 & 64.32 & \textbf{71.88} & 74.34 & 62.25 & \textbf{73.13} & \textbf{72.64} & 78.73 & 70.34 \\
        \rowcolor{gray!10} +Ours & \textbf{63.26} & \textbf{81.11} & \textbf{60.39} & \textbf{67.38} & \textbf{82.99} & \textbf{67.23} & \textbf{71.88} & \textbf{74.83} & \textbf{64.40} & 71.88 & 71.56 & \textbf{82.67} & \textbf{71.63} \\
        \rowcolor{gray!10} & 0.00 & \textcolor{mblue}{+4.01} & \textcolor{mblue}{+0.63} & \textcolor{mblue}{+0.69} & \textcolor{mblue}{+2.97} & \textcolor{mblue}{+2.91} & 0.00 & \textcolor{mblue}{+0.49} & \textcolor{mblue}{+2.15} & -1.25 & -1.08 & \textcolor{mblue}{+3.94} & \textcolor{mblue}{+1.29} \\
        \bottomrule
    \end{tabular}
    \caption{\textbf{Comparison of accuracy on four domains of DomainNet}: \textbf{C}: Clipart, \textbf{P}: Painting, \textbf{R}: Real, \textbf{S}: Sketch.}
    \vspace{-0.3cm}
    \label{Table:experiment-domainnet}
\end{table*}





    




\noindent\textbf{VisDA-C~\cite{peng2017visda}.} 
VisDA-C is a challenging large-scale benchmark whose training data is synthesized through 3D model rendering, and its test data is sampled from the real world.
The dataset contains 12 categories.
We utilize an imbalanced dataset VisDA-C (RSUT), where source and target domains are subject to two reverse Pareto distributions, following the previous work~\cite{tan2020class}.
Here, RSUT denotes the combination of Reversely-unbalanced Source (RS) and Unbalanced Target (UT) distribution.
The label distributions of RSUT are described in the supplementary.

\noindent\textbf{OfficeHome~\cite{venkateswara2017deep}.} 
This dataset comprises four domains, each consisting of 65 categories.
We also employ OfficeHome (RSUT)~\cite{tan2020class}, which is created by the same protocol as VisDA-C (RSUT).
Since the artistic domain in OfficeHome is too small to sample an imbalanced subset, we only utilize the remaining three distinct domains (\textit{e.g.}, Clip Art, Product, and Real-World), as prior work~\cite{li2021imbalanced}.

\noindent\textbf{DomainNet~\cite{peng2019moment}.}
We employ a subset of DomainNet~\cite{tan2020class}, comprising 40 categories from four domains: Real, Clipart, Painting, and Sketch.
As the label shift between these domains is inherent, we did not need to modify the label distribution of the dataset.
The visualization of label distribution in DomainNet are illustrated in the supplementary.


\noindent\textbf{Baseline Methods.}
Note that we utilize the pre-trained model trained using \emph{balanced softmax}~\cite{ren2020balanced}, which is a widely used long-tailed recognition approach.
\textbf{Source} indicates the pre-trained model with the source data using balanced softmax.
We compare our method with the following TTA baselines: BN stats~\cite{schneider2020improving}, ONDA~\cite{mancini2018kitting}, PseudoLabel~\cite{lee2013pseudo}, LAME~\cite{boudiaf2022parameter}, CoTTA~\cite{wang2022continual}, TENT~\cite{wang2020tent}, IABN~\cite{gong2022note}, and NOTE~\cite{gong2022note}.
Note that IABN is a normalization layer introduced in the NOTE paper.
Since NOTE has been proposed for temporally correlated test samples, IABN layer has the capability to handle the class-imbalance in a batch.

\begin{table}[t!]
    \scriptsize
    \centering
    \begin{tabular}{@{}l@{}c@{ }c@{\quad}c@{\quad}c@{\quad}c@{\quad}c@{\quad}c@{\quad}c@{ }c@{}}
        \toprule
        \multirowcell{2}{\textbf{Dataset}} & \multirowcell{2}{\textbf{Prior}} & \multicolumn{3}{c}{\textbf{Forward-LT}} & \textbf{Uni.} & \multicolumn{3}{c}{\textbf{Backward-LT}} & \multirowcell{2}{\multicolumn{1}{c}{\textbf{Avg.}}} \\
        \cmidrule(lr){3-5} \cmidrule(lr){6-6} \cmidrule(lr){7-9}
        && 50 & 25 & 10 & 1 & 10 & 25 & 50 \\
        \midrule
        \multirowcell{2}{CIFAR-10-C}
        & \xmark & 80.58 & 78.62 & \textbf{75.26} & 63.34 & 68.54 & 70.07 & 71.64 & 72.58 \\
        & \cmark & \textbf{81.58} & \textbf{79.05} & 75.17 & \textbf{69.55} & \textbf{68.65} & \textbf{70.51} & \textbf{72.86} & \textbf{73.91} \\
        \midrule
        \multirowcell{2}{CIFAR-100-C}
        & \xmark & 52.06 & 49.71 & 46.03 & 36.84 & 29.29 & 26.33 & 25.50 & 37.97 \\
        & \cmark & \textbf{53.59} & \textbf{50.95} & \textbf{46.96} & \textbf{37.17} & \textbf{29.79} & \textbf{27.46} & \textbf{27.03} & \textbf{38.99} \\
        \midrule
        \multirowcell{2}{ImageNet-C}
        & \xmark & 46.88 & 45.16 & 42.68 & 35.72 & \textbf{33.18} & \textbf{32.91} & 33.17 & 38.53 \\
        & \cmark & \textbf{47.71} & \textbf{45.69} & \textbf{42.98} & \textbf{36.02} & 32.95 & 32.65 & \textbf{33.23} & \textbf{38.75} \\
        \bottomrule
    \end{tabular}
    \caption{\textbf{Comparison between estimated label distribution and target prior on CIFAR-10-C, CIFAR-100-C, and ImageNet-C.} Prior indicates that the true target label distribution is utilized as prior knowledge for label shift adapter, instead of estimated label distribution. We conduct the experiments using IABN+Ours~\cite{gong2022note}.}
    \vspace{-0.3cm}
    \label{Table:experiment-gt-prior}
\end{table}

\begin{table*}[t!]
    \parbox{0.71\textwidth}{
        \small
        \centering
        \begin{tabular}{@{}llc@{\quad}c@{\quad}c@{\quad}c@{\quad}c@{\quad}c@{\quad}c@{\quad}c@{}}
            \toprule
            \multirowcell{2}{\textbf{Dataset}} & \multirowcell{2}{\textbf{Method}} & \multicolumn{3}{c}{\textbf{Forward-LT}} & \textbf{Uni.} & \multicolumn{3}{c}{\textbf{Backward-LT}} & \multirowcell{2}{\multicolumn{1}{c}{\textbf{Avg.}}} \\
            \cmidrule(lr){3-5} \cmidrule(lr){6-6} \cmidrule(lr){7-9}
            && 50 & 25 & 10 & 1 & 10 & 25 & 50 \\
            \midrule
            \multirowcell{3}{CIFAR-10-C}
            & Logit Adjust & 78.89 & 77.20 & 73.58 & 58.86 & 46.12 & 41.95 & 39.72 & 59.47 \\
            & IM Loss & 76.54 & 75.39 & 72.34 & 56.43 & 49.79 & 47.58 & 46.62 & 60.67 \\
            & Ours+IABN & \textbf{80.58} & \textbf{78.62} & \textbf{75.26} & \textbf{63.34} & \textbf{68.54} & \textbf{70.07} & \textbf{71.64} & \textbf{72.58} \\
            \midrule
            \multirowcell{3}{CIFAR-100-C}
            & Logit Adjust & 43.23 & 41.16 & 37.55 & 30.94 & 25.44 & 23.67 & 23.38 & 32.19 \\
            & IM Loss & 44.10 & 42.64 & 40.00 & 33.36 & 23.61 & 20.54 & 19.11 & 31.91 \\
            & Ours+IABN & \textbf{52.06} & \textbf{49.71} & \textbf{46.03} & \textbf{36.84} & \textbf{29.29} & \textbf{26.33} & \textbf{25.50} & \textbf{37.97} \\
            \bottomrule
        \end{tabular}
        \caption{Comparison with logit adjustment~\cite{menon2020long,hong2021disentangling} and information maximization (IM) loss~\cite{liang2020we}.}
        \label{Table:experiment-simple-baseline}
    }\hfill
    \parbox{0.27\textwidth}{
        \small
        \centering
        \begin{tabular}{@{}lcc@{}}
            \toprule
             \textbf{Method} & \textbf{MACs} & \textbf{Params} \\
            \midrule
             ResNet-18 & 557.93M & 11.22M \\
             \rowcolor{gray!10} + Ours & 558.04M & 11.34M \\
            \bottomrule
        \end{tabular}
        \caption{\textbf{Computational costs.} We measure MACs and the number of parameters (Params.).}
        \label{Table:experiment-cost}
    }
    \vspace{-0.6cm}
\end{table*}

\noindent\textbf{Implementation Details.}
We utilize ResNet-18~\cite{he2016deep} as the backbone for CIFAR-10 and CIFAR-100, and ResNeXt~\cite{xie2017aggregated} for ImageNet.
We also employ ResNet-50 for OfficeHome and DomainNet, and ResNet-101 for VisDA-C, which are pre-trained on ImageNet.
For fair comparisons, the same architecture and optimizer are utilized for all TTA baselines.
In all experiments, the source domain model is trained using Balanced softmax~\cite{ren2020balanced}, a representative long-tailed recognition method.
During inference, the batch size is set to 64 in all experiments.
Further details regarding the hyperparameters for each baseline and our method are described in the supplementary.

\subsection{Results on Corruption Data}

Table~\ref{Table:experiment-cifar} reports the average accuracy on 15 corruption types in CIFAR-10-C and CIFAR-100-C.
The results on each target domain are presented in the supplementary.
The results demonstrate that the performance of previous TTA methods is significantly inferior in the backward long-tailed distributions compared to the forward settings.
This issue arises because TTA models learn based on model predictions, which are biased toward the majority classes.

In contrast, it is noteworthy that our strategy considerably improves the accuracy when label shift is severe.
As our method produces the optimal parameters depending on the target label distribution, the models can be adapted to the target domain stably, even in the presence of severe label shifts.
Furthermore, when integrated with IABN~\cite{gong2022note}, a normalization layer for addressing the class imbalance in a batch, our method yields the best performance.
Nevertheless, it should be noted that relying solely on IABN may not be sufficient in managing severe label shifts.

As shown in Table~\ref{Table:experiment-imagenet}, our method consistently outperforms the existing TTA approaches on ImageNet-C, a more challenging dataset.
Integrating our proposed method consistently improves the performance in all test sets.
Furthermore, our method also improves the accuracy on the uniform dataset: TENT (+5.14\%) and IABN (+0.52\%).
The promising results demonstrate the practicality and effectiveness of our method under covariate and label shifts.


\subsection{Results on DA Benchmarks}

We evaluate the effectiveness of our method in comparison with TTA methods on three domain adaptation benchmarks: VisDA-C (RSUT), OfficeHome (RSUT), and DomainNet.
In VisDA-C (RSUT) and OfficeHome (RSUT), we utilize the test datasets, including the reversed Pareto label distribution.
Surprisingly, the existing TTA methods perform worse than the source pre-trained model for evaluation data of VisDA-C, as shown in Table~\ref{Table:experiment-visda}.
This result indicates that the baselines do not work when the label shift is severe.
In contrast, TENT combined with our method improves the performance significantly for VisDA-C test data.

The results on OfficeHome (RSUT) are reported in Table~\ref{Table:experiment-officehome}.
We discovered that since the number of test samples in OfficeHome (RSUT) is limited (\textit{e.g.}, Clipart: 1,017, Product: 1,985, Real: 1,235), TTA methods generally do not result in significant performance improvements compared to the source model.
Nonetheless, our method exhibits a general improvement on the OfficeHome (RSUT) datasets, except for P$\rightarrow$C.

Table~\ref{Table:experiment-domainnet} shows the results on the DomainNet dataset, where the label shifts between different domains already exist.
This result demonstrates that our approach generally outperforms the baselines.
Moreover, we confirmed that our model performs better when adapting to Real domain that contains a larger number of test samples compared to other domains (\textit{e.g.}, Real: 6,943, Clipart: 1,616, Painting: 2,909, Sketch: 2,399).
It is because the target label distribution can be estimated more precisely with more test samples.

\subsection{Analysis on Label Shift Adapter}

\begin{figure}[t!]
    \centering
    \includegraphics[width=\linewidth]{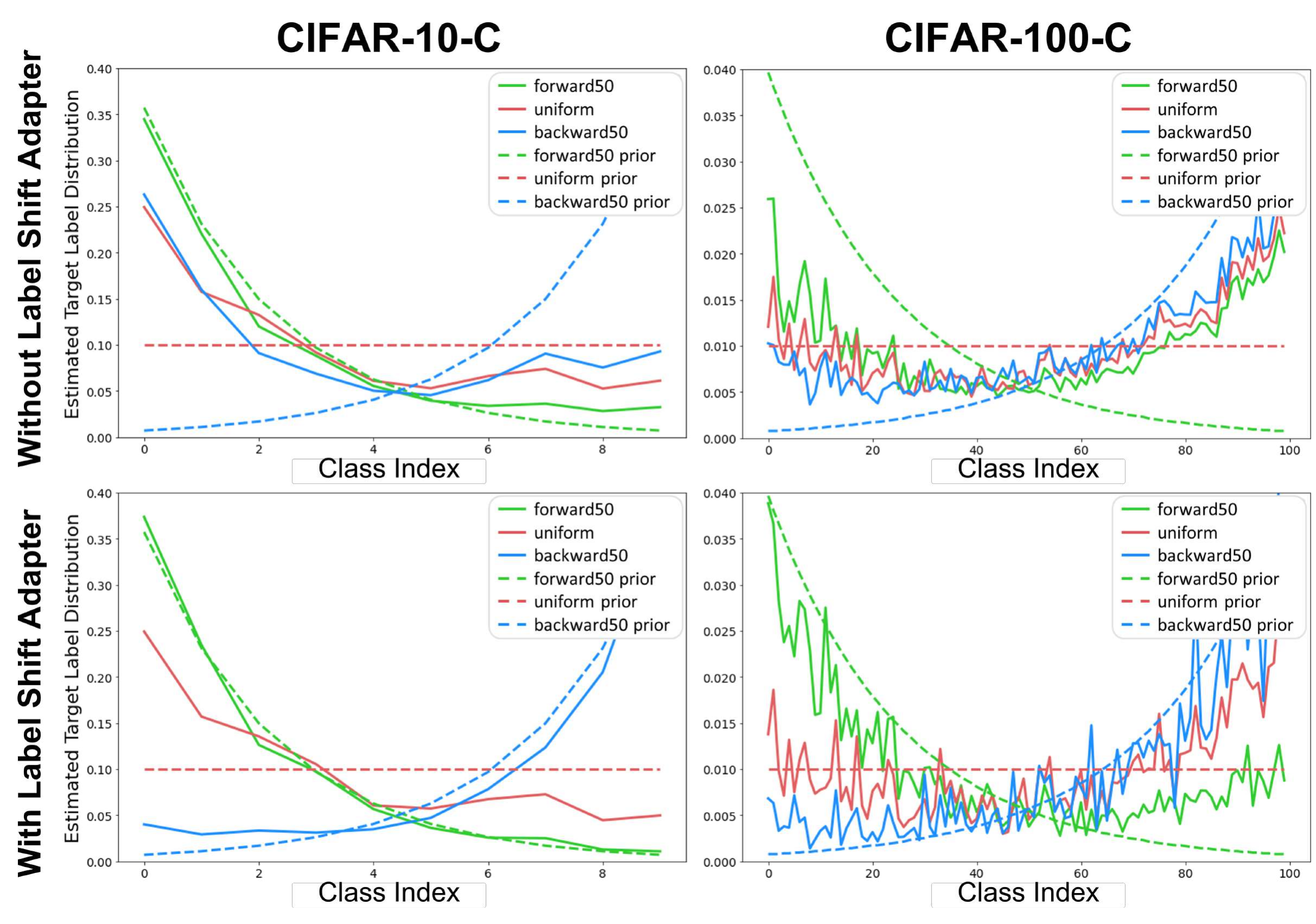}
    \vspace{-0.3cm}
    \caption{Visualization of estimated target label distributions and target priors on CIFAR-10-C and CIFAR-100-C. Applying the label shift adapter improves the target label distribution estimation.}
    \vspace{-0.5cm}
    \label{fig:estimated-distribution}
\end{figure}

\noindent\textbf{Estimated Target Label Distribution.}
The proposed label shift adapter predicts the weight difference with the estimated target label distribution.
Therefore, it is essential to estimate the target class prior accurately, which is the label distribution of target domain data.
Different from previous TTA methods, our method adapts the pre-trained model according to the label distribution shift.
As shown in Fig.~\ref{fig:estimated-distribution}, we visualize the estimated target label distribution of the model with and without our label shift adapter.
This result shows that the model with the label shift adapter is superior for estimating various label distributions compared to the model without the label shift adapter.
In addition, our method estimates the target label distribution similar to the target prior in CIFAR-10-C and CIFAR-100-C datasets.

As shown in Table~\ref{Table:experiment-gt-prior}, we compare the performances of the label shift adapter with estimated target label distribution and target prior.
Obviously, the overall accuracy of the model increases if the target prior is utilized instead of the estimated target label distribution.
Nevertheless, our method shows comparable performances to the model using target prior as the input of label shift adapter.

\noindent\textbf{Computational Costs.}
Table~\ref{Table:experiment-cost} shows the computational costs of our method.
We measure the computational cost of ResNet-18, which is utilized for CIFAR-100-C.
Since we design the architecture of the label shift adapter efficiently, our method requires a negligible amount of additional computational costs (MACs: +0.11M, Params: +0.12M).
We describe the details of the label shift adapter architecture in the supplementary.
Moreover, we report ablation study on the architecture of label shift adapter in the supplementary.

\noindent\textbf{Effectiveness of Label Shift Adapter.}
We validate the effectiveness of our method over existing approaches for handling label shifts.
Specifically, we apply the following non-trivial baselines with entropy minimization loss and IABN~\cite{gong2022note}: (i) Post-hoc logit adjustment~\cite{menon2020long,hong2021disentangling} modifies the logits using the estimated target label distribution. (ii) Information maximization loss~\cite{liang2020we} makes the outputs globally diverse by maximizing mean entropy, which can reduce the bias toward certain classes.
Table~\ref{Table:experiment-simple-baseline} shows that these baselines are not sufficient to handle the label distribution shifts, particularly on inversely long-tailed distribution (\textit{e.g.}, Backward).
In contrast, our approach leads to promising performance gains on various label distributions.

\section{Discussions}

\noindent\textbf{Normalization Layers for Label Shifts.}
This paper introduces the label shift adapter for addressing the label shift problems in the test-time adaptation scenario, which can be applied to any model regardless of its architecture.
In addition to our method, we found that managing the bias in batch statistics is also crucial for reducing performance degradation caused by label shifts.
This observation is evidenced by the effectiveness of normalization layers, such as instance-aware batch normalization (IABN) layers, in dealing with the class imbalance in a batch.
However, IABN layers are highly sensitive to hyperparameter selections such as the soft-shrinkage width $\alpha$.
Consequently, improving the robustness of normalization layers to label shifts is a promising future research direction.

\noindent\textbf{Training Additional Component.}
One limitation of our work is that our method requires an additional stage for training the label shift adapter.
In several recent test-time adaptation methods~\cite{choi2022improving,zou2022learning,lim2023ttn}, an additional training process is often carried out before server deployment to train the additional components.
Despite the drawback of requiring an additional training stage, we believe that these test-time adaptation models, including our method, are practically useful because they are more robust during inference.
Furthermore, there are techniques~\cite{kang2019decoupling,cao2019learning,xiang2020learning,wang2020long,alshammari2022long} in the long-tailed recognition field that involves dividing the training process into two stages.
Our approach can serve as an inspiration for methods that can be applied for handling test-agnostic label distributions in long-tailed recognition, regardless of the model architecture.

\section{Conclusion}

This paper addresses the label shift problem in TTA, where both source and target domain datasets are class imbalanced.
Existing TTA methods employing entropy minimization are often flawed due to the model bias toward the majority classes in source data.
To address such a non-trivial issue, we propose a novel label shift adapter, which produces the optimal parameters according to the label distribution.
Our label shift adapter is applicable to existing TTA methods regardless of the model architectures.
Furthermore, we estimate the label distribution of target domain data to feed into the label shift adapter.
Through extensive experiments, we demonstrate that our method outperforms the state-of-the-art TTA baselines.
We believe that our work inspires future researchers to improve TTA methods under the joint presence of covariate and label shifts.

\noindent\textbf{Acknowledgements.} 
We would like to thank Kyuwoong Hwang, Simyung Chang, Hyunsin Park, Janghoon Cho, Juntae Lee, Hyoungwoo Park, Seokeon Choi, and Jungsoo Lee of Qualcomm AI Research team for their valuable discussions.

{\small
\bibliographystyle{ieee_fullname}
\bibliography{reference}
}

\clearpage

\begin{figure}
\twocolumn[{
\renewcommand\twocolumn[1][]{#1}
    \centering 
    \vspace*{1.1cm}
    \Large{\bf{Supplementary Material}}
    \vspace*{1.7cm}
    
    \includegraphics[width=\linewidth]{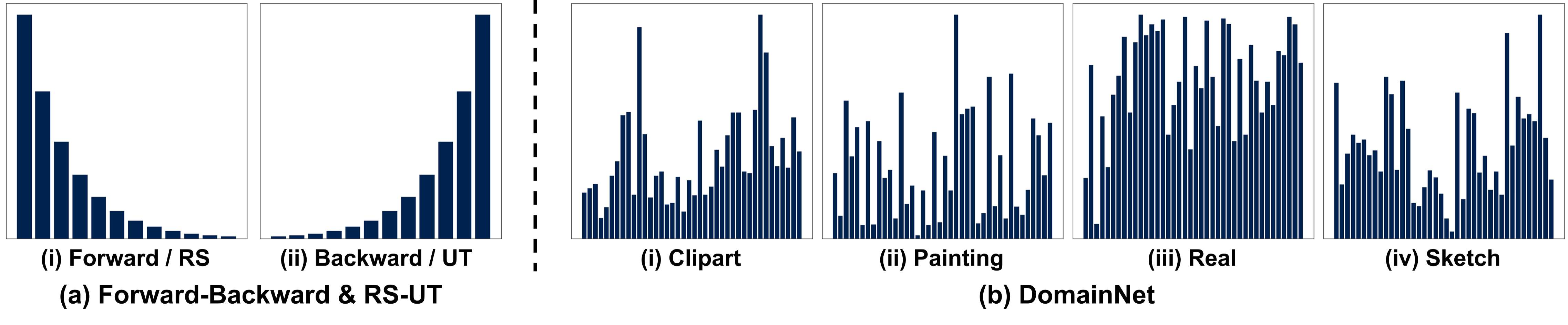}
    \vspace{-0.6cm}
    \caption{\textbf{Visualization of label distributions in datasets.} (a) shows the illustrations of forward or reversely-unbalanced source (RS) setting and backward or unbalanced target (UT) setting. In specific, forward and backward are used in CIFAR-10-C, CIFAR-100-C, and ImageNet-C. In addition, RS and UT are utilized in VisDA-C and OfficeHome. (b) shows the natural label shift of DomainNet.}
    \vspace{0.2cm}
    \label{supp-fig:data-distribution}
}]
\end{figure}

\begin{table*}[t!]
    \small
    \centering
    \begin{tabular}{cccccccccc}
        \toprule
        \textbf{Src Data} & \textbf{Tgt Data} & \textbf{Model} & \textbf{Optim.} & \textbf{Scheduler} & \textbf{Epoch} & \textbf{Batch} & \textbf{WD} & \textbf{Momentum} & \textbf{LR} \\
        \midrule
        CIFAR-10-LT & CIFAR-10-C & ResNet-18 & SGD & CosineAnneal & 200 & 128 & 5e-4 & 0.9 & 0.1 \\
        CIFAR-100-LT & CIFAR-100-C & ResNet-18 & SGD & CosineAnneal & 200 & 128 & 5e-4 & 0.9 & 0.1 \\
        ImageNet-LT & ImageNet-C & ResNeXt-50 & SGD & Manual & 90 & 64 & 2e-4 & 0.9 & 0.1 \\
        VisDA-C (RS) & VisDA-C (UT) & ResNet-101 & SGD & - & 15 & 40 & 1e-3 & 0.9 & 1e-3 \\
        OfficeHome (RS) & OfficeHome (UT) & ResNet-50 & SGD & - & 50 & 40 & 1e-3 & 0.9 & 1e-2 \\
        DomainNet & DomainNet & ResNet-50 & SGD & - & 20 & 40 & 1e-3 & 0.9 & 1e-2 \\
        \bottomrule
    \end{tabular}
    \caption{\textbf{Hyperparameters for training the model with source domain data.} Src Data and Tgt Data denote source domain and target domain datasets, respectively. Optim. indicates the optimizer. WD and LR denote the weight decay and learning rate for training. The manual scheduler for ImageNet-LT is to decay the learning rate at 60 and 80 epochs.}
    \label{Table:hyperparameters}
\end{table*}

\begin{table*}[t!]
    \small
    \centering
    \begin{tabular}{p{0.04\textwidth}<{\centering}p{0.04\textwidth}<{\centering}p{0.04\textwidth}<{\centering}p{0.04\textwidth}<{\centering}|p{0.05\textwidth}<{\centering}p{0.05\textwidth}<{\centering}p{0.05\textwidth}<{\centering}p{0.05\textwidth}<{\centering}p{0.05\textwidth}<{\centering}p{0.05\textwidth}<{\centering}p{0.05\textwidth}<{\centering}p{0.05\textwidth}<{\centering}p{0.05\textwidth}<{\centering}p{0.05\textwidth}<{\centering}p{0.05\textwidth}<{\centering}}
        \toprule
        \multicolumn{4}{c}{\textbf{Model Architecture}} & \multicolumn{3}{c}{\textbf{Forward-LT}} & \textbf{Uni.} & \multicolumn{3}{c}{\textbf{Backward-LT}} & \multirowcell{2}{\textbf{Avg.}} \\
        \cmidrule(lr){1-4} \cmidrule(lr){5-7} \cmidrule(lr){8-8} \cmidrule(lr){9-11}
        $\gamma_h$ & $\beta_h$ & $\Delta W$ & $\Delta b$ & 50 & 25 & 10 & 1 & 10 & 25 & 50 \\
        \midrule
        \cmark &  &  &  & 51.20 & 49.30 & 46.06 & 37.36 & 27.28 & 23.75 & 21.84 & 36.69 \\
         & \cmark &  &  & 48.79 & 42.45 & 32.10 & 15.21 & 22.52 & 24.09 & 25.14 & 30.04 \\
         &  & \cmark &  & 51.32 & 49.36 & 46.11 & \underline{37.17} & 27.06 & 23.56 & 21.71 & 36.61 \\
         &  &  & \cmark & \textbf{52.50} & \textbf{50.19} & \textbf{46.64} & \textbf{37.51} & 28.95 & 25.56 & 24.06 & 37.92 \\
        \cmark & \cmark &  &  & \underline{52.09} & 49.48 & 45.43 & 35.92 & \textbf{29.60} & \textbf{26.82} & \textbf{26.25} & 37.94 \\
         &  & \cmark & \cmark & 51.93 & \underline{49.78} & \underline{46.43} & 37.18 & 27.71 & 24.28 & 22.57 & 37.12 \\
        \cmark & \cmark & \cmark & \cmark & 52.06 & 49.71 & 46.03 & 36.84 & \underline{29.29} & \underline{26.33} & \underline{25.50} & \textbf{37.97} \\     
        \bottomrule
    \end{tabular}
    \caption{\textbf{Ablation study on architecture design of label shift adapter using CIFAR-100-LT and CIFAR-100-C.}}
    \label{Table:ablation study}
\end{table*}

\section*{A. Implementation Details}

In this section, we introduce further information regarding the datasets, along with the implementation details for the baseline test-time-adaptation (TTA) methods and the label shift adapter.

\subsection*{A.1. Datasets}\label{A.1. Datasets}

Fig.~\ref{supp-fig:data-distribution} illustrates the label distributions for the datasets utilized in our experiments.
As depicted in Fig.~\ref{supp-fig:data-distribution} (a), `forward' and `RS' represent long-tailed label distributions, with class order corresponding to the training label distribution.
Conversely, `backward' and `UT' indicate a reversed class order.

In the forward and backward settings, the imbalance ratios for CIFAR-10-C, CIFAR-100-C, and ImageNet-C are configured to 10, 25, and 100.
We adjust the label distribution by reducing the number of images per class based on the specified imbalance ratio.
For VisDA-C, The imbalance ratio is set to 100 for both training and test datasets.
Furthermore, we utilize an imbalanced version of OfficeHome created by the previous research~\cite{tan2020class}.

Fig.~\ref{supp-fig:data-distribution} (b) shows the label distributions of DomainNet, in which existing label shifts are significant enough.
The superior performance of our method on DomainNet demonstrates its ability to handle label shifts that arise in real-world scenarios.

\subsection*{A.2. Details of Baselines}

We carry out the experiments using the official implementations of the baseline models.
We provide additional details regarding the implementation specifics, including hyperparameters.
Note that the batch size for test-time adaptation is configured to 64 for fair comparisons.
For simplicity, we present the hyperparameters in the following sequence: \textbf{\{CIFAR-10-C, CIFAR-100-C, ImageNet-C, VisDA-C, OfficeHome, DomainNet\}} for test-time adaptation baselines.
In instances where hyperparameters are not separately described for each dataset, the same values are employed across all datasets.

\noindent\textbf{Source.}
Different from the previous TTA studies, we employ long-tailed datasets in our research.
To mitigate model bias towards the majority classes, we utilize a balanced softmax~\cite{ren2020balanced}, which is a prominent method for long-tailed recognition.
Formally, the balanced softmax is expressed as:
\begin{equation*}
  \mathcal{L}_{\text{bal}} = - \sum_{(x_i, y_i) \sim \mathcal{D}_s} y_i \log \sigma (\hat{y}_i + \log(\pi_s)),
\end{equation*}
where $\pi_s$ represents the frequency of the training classes, and $\sigma$ denotes the softmax function.

Table~\ref{Table:hyperparameters} describes the hyperparameters utilized for training on source domain datasets.
We select the hyperparameters for VisDA-C, OfficeHome, and DomainNet in accordance with the imbalanced source-free domain adaptation study~\cite{li2021imbalanced}.
As described in the main manuscript, we utilize pre-trained ResNet-50 and ResNet-101 on ImageNet, when conducting the experiments on VisDA-C, OfficeHome, and DomainNet.
Moreover, the learning rate for the feature extractor and the classifier is set to 0.1$\times$LR and LR, respectively, when training the model on VisDA-C, OfficeHome, and DomainNet.
All experiments are conducted using NVIDIA RTX A5000 GPU.

\noindent\textbf{BN Stats.}
BN stats~\cite{schneider2020improving} utilizes test batch statistics instead of running statistics within batch normalization layers.

\noindent\textbf{PseudoLabel.}
In accordance with previous studies~\cite{lee2013pseudo,wang2020tent}, 
we update the affine parameters in the batch normalization layers using the hard pseudo labels.
The learning rate is set to \{1e-3, 1e-3, 2.5e-4, 5e-5, 5e-5, 1e-3\} for each respective dataset, following the hyperparameters of TENT~\cite{wang2020tent}.

\noindent\textbf{ONDA.}
Online domain adaptation (ONDA)~\cite{mancini2018kitting} modifies the batch normalization statistics for target domains using a batch of target data through an exponential moving average.
We set the update frequency $N=10$ and the decay of the moving average $m=0.1$, adhering to the default values of the original paper.

\noindent\textbf{TENT.}
Test entropy minimization (TENT)~\cite{wang2020tent} optimizes the affine parameters of batch normalization layers via entropy minimization.
The learning rate is configured to \{1e-3, 1e-3, 2.5e-4, 5e-5, 5e-5, 1e-3\} for each dataset.
We referred to the official implementation for hyperparameter selection.

\noindent\textbf{LAME.}
Laplacian adjusted maximum-likelihood estimation (LAME)~\cite{boudiaf2022parameter} alters the output probability of the classifier.
Following the authors' implementation, we set the kNN affinity matrix with the value of $k$ as 5.

\noindent\textbf{CoTTA.}
Continual test-time adaptation (CoTTA)~\cite{wang2022continual} adapts the model to accommodate continually evolving target domains by employing a weight-averaged teacher model, data augmentations, and stochastic restoring.
CoTTA incorporates three hyperparameters: augmentation confidence threshold $p_{th}$, restoration factor $p$, and the decay of EMA $m$.
$p$ and $m$ are set to 0.01 and 0.999, respectively.
Additionally, $p_{th}$ is configured to \{ 0.92, 0.72, 0.01, 0.01, 0.01, 0.01 \}.
Given that the authors do not provide the hyperparameters for VisDA-C, OfficeHome, and DomainNet, we fine-tune the appropriate hyperparameters for them.

\noindent\textbf{NOTE.}
Non-i.i.d. test-time adaptation (NOTE)~\cite{gong2022note} comprises two components: (i) Instance-aware batch normalization (IABN), and (ii) Prediction-balanced reservoir sampling (PBRS).
In accordance with the original paper, we substitute the batch normalization layers with IABN layers before pre-training the source models.
Two hyperparameters are associated with IABN: soft-shrinkage width $\alpha$ and EMA momentum $m$.
The values of $\alpha$ are configured as \{ 4, 4, 8, 8, 8, 8 \}, while $m$ is set to \{ 0.01, 0.01, 0.1, 0.1, 0.1, 0.1 \}.
The memory size of PBRS is set to 64, equal to the batch size.
In our experiments, we incorporate our label shift adapter into the models using IABN layers.

\subsection*{A.3. Details of Label Shift Adapter}

\begin{algorithm}[t!]
\caption{Training Process of Label Shift Adapter} \label{alg:model}
\begin{algorithmic}[1]

\Require Dataset $\mathcal{D}_s = \{(x_i,y_i)\}_{i=1}^n$. A pre-trained model $f$. A label shift adapter $\mathcal{G}_{\phi}$.

\State Initialize the parameters $\phi$ randomly
\For {$k=1$ to $K$}
	\State  $\mathcal{B}\leftarrow \text{SampleMiniBatch}(\mathcal{D}, m)$
    \Statex \Comment{a mini-batch of $m$ examples}
    \State $\pi, \tau \leftarrow \text{Sample}(\{\pi_s, u, \bar{\pi}_s \}, \{ \tau_{\pi_s}, \tau_{u}, \tau_{\bar{\pi}_s} \})$ 
    \Statex \Comment{sample $\tau$ matching $\pi$}
    \State $\mathcal{L}(\mathcal{G}_{\phi}) \leftarrow \frac{1}{m} \sum_{(x,y)\in \mathcal{B}}\mathcal{L}_{gla}((x,y,\pi);f, \mathcal{G}_{\phi}) $ 
    \State $\mathcal{G}_{\phi} \leftarrow \mathcal{G}_{\phi} - \eta\nabla_\theta \mathcal{L}(\mathcal{G}_{\phi})$ \Comment one SGD step
\EndFor
\end{algorithmic}
\end{algorithm}

\begin{figure*}[t!]
    \centering
    \includegraphics[width=0.75\linewidth]{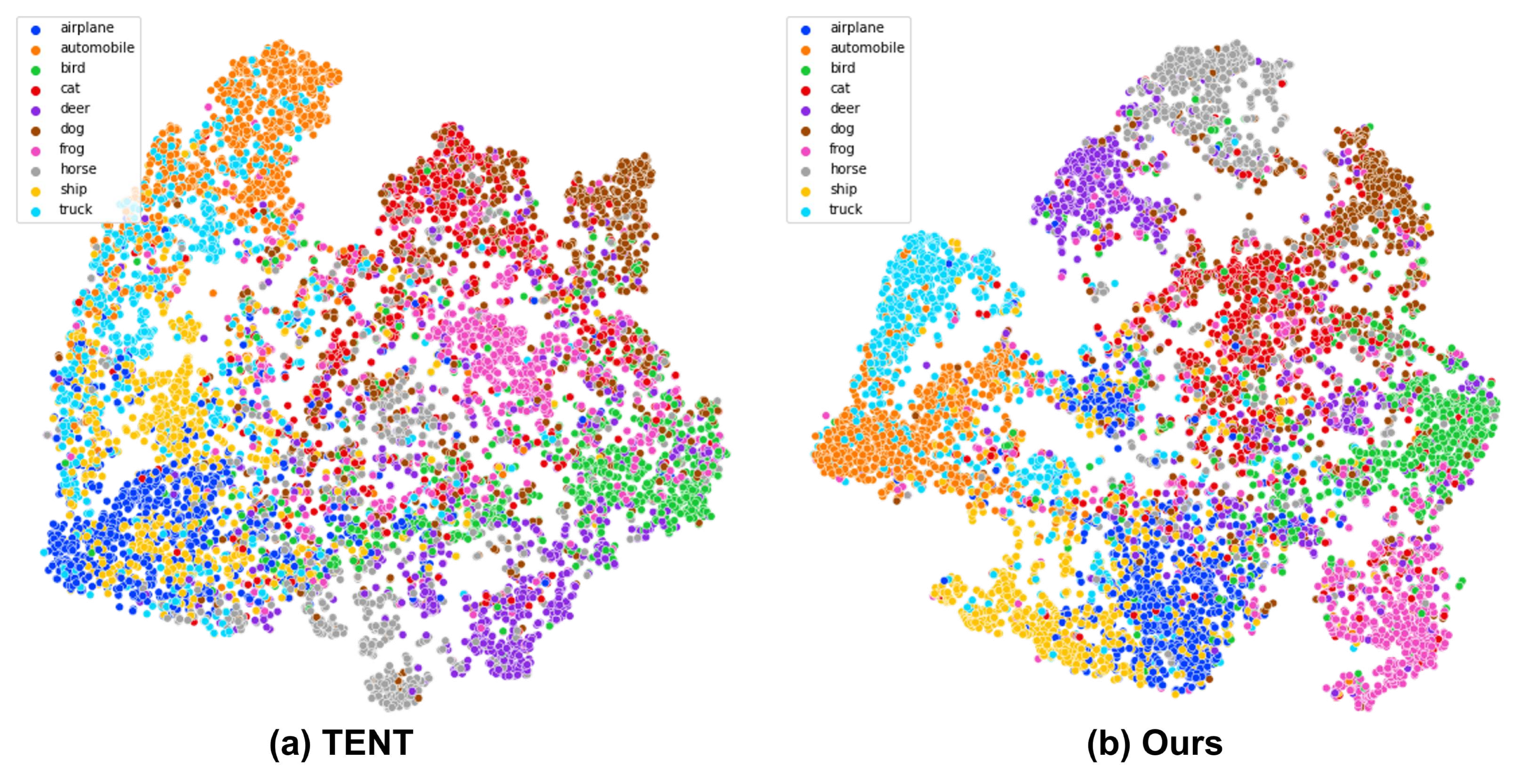}
    \caption{\textbf{T-SNE visualizations of (a) TENT and (b) IABN+Ours.} We visualize the feature map $h$ obtained from the Gaussian noise corruption in the CIFAR-10-C uniform test dataset. The number of training samples is large in the order of the classes in the legend.}
    \label{supp-fig:tsne}
\end{figure*}

\begin{table*}[t!]
    \small
    \centering
    \begin{tabular}{clcccccccc}
        \toprule
        \multirowcell{2}{\textbf{Src Method}} & \multirowcell{2}{\textbf{TTA Method}} & \multicolumn{3}{c}{\textbf{Forward-LT}} & \textbf{Uni.} & \multicolumn{3}{c}{\textbf{Backward-LT}} & \multirowcell{2}{\multicolumn{1}{c}{\textbf{Avg.}}} \\
        \cmidrule(lr){3-5} \cmidrule(lr){6-6} \cmidrule(lr){7-9}
        && 50 & 25 & 10 & 1 & 10 & 25 & 50 \\
        \midrule
        \multirowcell{10}{\textit{Cross} \\ \textit{Entropy}} & Source & 40.20 & 37.26 & 33.18 & 22.20 & 12.19 & 8.95 & 7.24 & 23.03 \\
        \cmidrule(lr){2-10}
        & BN Stats & 48.12 & 45.44 & 40.17 & 26.17 & 13.67 & 9.69 & 7.80 & 27.30 \\
        & ONDA & 48.49 & 45.80 & 41.16 & 27.66 & 15.17 & 11.02 & 8.98 & 28.33 \\
        & PseudoLabel & 48.76 & 45.26 & 39.09 & 19.84 & 11.32 & 8.19 & 6.44 & 25.56 \\
        & TENT & 49.17 & 45.21 & 38.42 & 15.32 & 9.99 & 7.44 & 5.67 & 24.46 \\
        & LAME & 38.17 & 35.17 & 31.22 & 20.44 & 11.02 & 7.93 & 6.30 & 21.46 \\
        & CoTTA & 32.83 & 29.35 & 25.72 & 14.75 & 7.38 & 4.94 & 3.67 & 16.95 \\
        & NOTE & 46.41 & 44.10 & 40.49 & 29.30 & 15.77 & 11.87 & 9.84 & 28.26 \\
        \cmidrule(lr){2-10}
        & IABN & 46.43 & 43.92 & 39.80 & 25.35 & 14.71 & 11.13 & 9.27 & 27.23 \\
        \rowcolor{gray!10} & +Ours & \textbf{53.20} & \textbf{50.77} & \textbf{46.26} & \textbf{32.34} & \textbf{18.56} & \textbf{14.07} & \textbf{11.74} & \textbf{32.42} \\
        \midrule
        \multirowcell{10}{\textit{Balanced} \\ \textit{Sampling}} & Source & 34.88 & 32.54 & 29.12 & 20.29 & 11.80 & 9.09 & 7.51 & 20.75 \\
        \cmidrule(lr){2-10}
        & BN Stats & 45.07 & 42.93 & 39.10 & 28.67 & 17.82 & 13.97 & 11.88 & 28.49 \\
        & ONDA & 44.79 & 42.60 & 39.01 & 29.20 & 18.62 & 14.65 & 12.70 & 28.80 \\
        & PseudoLabel & 47.08 & 44.66 & 40.45 & 26.98 & 17.31 & 13.59 & 11.23 & 28.76 \\
        & TENT & 48.28 & 45.64 & 41.07 & 24.04 & 16.88 & 13.50 & 11.30 & 28.67 \\
        & LAME & 32.88 & 30.44 & 27.08 & 18.48 & 10.48 & 7.89 & 6.39 & 19.09 \\
        & CoTTA & 30.38 & 28.68 & 25.58 & 17.32 & 10.74 & 7.96 & 6.49 & 18.16 \\
        & NOTE & 46.62 & 44.27 & 40.64 & 30.42 & 16.99 & 12.77 & 10.32 & 28.86 \\
        \cmidrule(lr){2-10}
        & IABN & 46.85 & 44.29 & 40.40 & 27.20 & 16.12 & 12.05 & 9.88 & 28.12 \\
        \rowcolor{gray!10} & +Ours & \textbf{51.32} & \textbf{49.18} & \textbf{44.99} & \textbf{31.92} & \textbf{19.64} & \textbf{15.11} & \textbf{13.00} & \textbf{32.16} \\
        \midrule
        \multirowcell{10}{\textit{Classifier} \\ \textit{Re-Training}} & Source & 40.17 & 37.47 & 33.59 & 23.23 & 13.42 & 10.18 & 8.51 & 23.80 \\
        \cmidrule(lr){2-10}
        & BN Stats & 48.33 & 45.60 & 40.82 & 27.49 & 15.13 & 11.16 & 9.11 & 28.23 \\
        & ONDA & 48.33 & 45.77 & 41.57 & 28.96 & 16.78 & 12.50 & 10.63 & 29.22 \\
        & PseudoLabel & 49.10 & 45.89 & 40.19 & 21.01 & 12.74 & 9.50 & 7.66 & 26.58 \\
        & TENT & 49.70 & 46.19 & 39.79 & 16.59 & 11.28 & 8.93 & 6.92 & 25.63 \\
        & LAME & 38.22 & 35.41 & 31.60 & 21.48 & 12.28 & 9.13 & 7.52 & 22.23 \\
        & CoTTA & 31.93 & 29.23 & 25.91 & 15.26 & 8.18 & 5.55 & 4.33 & 17.20 \\
        & NOTE & 45.96 & 44.04 & 41.14 & 31.68 & 18.52 & 14.66 & 12.67 & 29.81 \\
        \cmidrule(lr){2-10}
        & IABN & 46.11 & 44.04 & 40.53 & 27.61 & 17.28 & 13.71 & 11.94 & 28.75 \\
        \rowcolor{gray!10} & +Ours & \textbf{53.11} & \textbf{50.86} & \textbf{46.66} & \textbf{34.04} & \textbf{20.77} & \textbf{16.41} & \textbf{14.12} & \textbf{33.71} \\
        \bottomrule
    \end{tabular}
    \caption{\textbf{Ablation study on the source pre-trained model using CIFAR-100-LT and CIFAR-100-C.}}
    \vspace{-0.5cm}
    \label{Table:experiment-src-ablation}
\end{table*}

\noindent\textbf{Model Architecture.}
We utilize the same model architecture for the label shift adapter across all datasets.
The proposed label shift adapter consists of two fully-connected (FC) layers and a ReLU activation function, structured as FC-ReLU-FC.
Furthermore, the label shift adapter is partitioned into two neural networks producing ($\gamma_h$, $\beta_h$) and ($\Delta W$, $\Delta b$).
As described in the main manuscript, the label shift adapter takes $m^\intercal \pi \in \mathbb{R}^{1}$ and produces ($\gamma_h$, $\beta_h$) and ($\Delta W$, $\Delta b$) in each respective neural network.
The hidden layer size in the label shift adapter is configured to 100.

\noindent\textbf{Details of Label Shift Adapter.}
We provide the algorithm of the training process for the label shift adapter as a pseudo-code in Algorithm~\ref{alg:model}.
The primary objective of the label shift adapter is to learn the relationship between $\pi$ and adaptive parameters by selecting appropriate $\tau$ based on sampled $\pi$ within generalized logit adjusted loss~\cite{menon2020long,aimar2022balanced} function.
Increasing $\tau$ results in decision boundary shifting away from the minority class towards the majority class.
Consequently, instead of sampling batches differently based on $\pi$, we sample $\pi$ and $\tau$ iteratively, as described in Algorithm~\ref{alg:model}.
This enables the label shift adapter to optimize its parameters in accordance with input label distributions (\textit{e.g.}, $\pi$ and $\hat{\mathcal{Y}}_t$), thereby producing suitable parameter adjustments.

During the training of the label shift adapter, we sample the label distribution $\pi$ from three types of label distributions: $\{ \pi_s, u, \bar{\pi}_s \}$.
For each sampled label distribution, we select the appropriate $\tau \subset \{ \tau_{\pi_s}, \tau_{u}, \tau_{\bar{\pi}_s} \}$, with the hyperparameter $\tau$ corresponding to each $\pi$.
Different $\tau$ values are employed for each dataset.
We set $\tau$ to $\{ 1, -1.5, 3 \}$, $\{ 1, 0, -2 \}$, $\{ 1, 0, -2 \}$, $\{ 1, 0, -2 \}$, $\{ 1, -1, -3 \}$, and $\{ 1, 0, -2 \}$, for CIFAR-10-LT, CIFAR-100-LT, ImageNet-LT, VisDA-C, OfficeHome, and DomainNet, respectively.

The mapping vector $m$ maps the label distribution's vector to the scalar of the imbalance degree.
We set the range of $m$ from -1 to 1, with the values increasing proportionally to the data count rank of each class.
This technique enables the adapter to effectively utilize the degree of imbalance as an input, circumventing the challenges associated with complex label spaces encountered when using $\pi$ directly.

While training the label shift adapter, we employ the same optimizer and batch size as those employed for training the source models.
The learning rate is set to 1e-3 for all datasets.
Moreover, we train the label shift adapter for $\{ 200, 200, 30, 15, 50, 20 \}$ epochs.

During inference, the momentum hyperparameter $\alpha$ for target label distribution estimation is configured to 0.1.
For learnable parameters in the test-time adaptation process, we only update affine parameters in normalization layers by following TENT~\cite{wang2020tent} and IABN~\cite{gong2022note}.
Unlike TENT, we freeze the top layers and update the affine parameters of the layer in the remaining shallow layers, inspired by previous work~\cite{choi2022improving,niu2023towards}.
Specifically, for ResNet, including four layer groups (layer 1, 2, 3, 4), we only freeze layer4 in CIFAR-10-C, CIFAR-100-C, and ImageNet-C.
In other datasets, there is no significant difference in performance, so all affine parameters are trained.
When estimating the label distribution on ImageNet-C, we utilize only the top-3 probability to update the estimated label distribution $\hat{\mathcal{Y}}_t$.
Empirically, we discovered that it is effective to consider only top-$k$ when the number of classes is particularly large.

\section*{B. Further Analysis on Label Shift Adapter}

\noindent\textbf{Ablation Study on Architecture Design.}
We examine the model architecture design for the proposed label shift adapter.
The label shift adapter produces four types of outputs: $\gamma_h$, $\beta_h$, $\Delta W$, and $\Delta b$.
Table~\ref{Table:ablation study} presents the ablation study for each component.
Interestingly, even when only $\Delta b$ is employed, the performance is quite good.
However, we observed that as the degree of the label shift increases, the performance of using only $\Delta b$ declines.
Moreover, utilizing $\gamma_h$ and $\beta_h$ only also yields impressive results, indicating that appropriately shiting the feature map $h$ is effective in addressing the label shifts.
We choose the architecture design of the label shift adapter that achieves the best average accuracy, indicating that the final model generally performs well across a variety of label distributions.

\noindent\textbf{T-SNE Visualization.}
To further substantiate the effectiveness of our method, we visualize the feature map $h$ using t-SNE by extracting $h$ during test-time adaptation.
As illustrated in Fig.~\ref{supp-fig:tsne}, our method shows a more well-separated representation space in a class-wise manner compared to TENT.
Notably, it is evident that the minority classes (\textit{e.g.}, horse and truck) are not well divided in the representation space of TENT.
In contrast, our method integrating into IABN layers enhances class-discriminability.

\noindent\textbf{Ablation Study on Source Model.}
In the main manuscript, we employ the balanced softmax to reduce the model bias towards the majority classes.
To further validate the effectiveness of the proposed method, we apply our method to several source pre-trained models utilizing different training strategies.
We employ three types of techniques: (i) Cross-entropy loss, (ii) Balanced sampling, (iii) Classifier re-training~\cite{kang2019decoupling}, where the feature extractor is trained using cross-entropy loss, and then the classifier is randomly re-initialized and re-trained using class-balanced sampling.
Table~\ref{Table:experiment-src-ablation} demonstrates that our method effectively handles the label shifts, regardless of the source pre-trained models.
Moreover, these results indicate that existing long-tailed recognition methods can be combined with our method to further reduce the model bias towards the majority classes in source domain data.

\begin{table}[t!]
    \centering
    \small
    \setlength{\tabcolsep}{3pt}
    \begin{tabular}{c|cccccccc}
        \toprule
        Num. & F50 & F25 & F10 & U & B10 & B25 & B50 & Avg. \\
        \midrule
        3 & \textbf{52.06} & \textbf{49.71} & \textbf{46.03} & 36.84 & \textbf{29.29} & 26.33 & 25.50 & \textbf{37.97} \\
        5 & 50.91 & 48.82 & 45.41 & 36.90 & 29.28 & \textbf{26.37} & \textbf{25.51} & 37.60 \\
        7 & 51.13 & 48.99 & 45.52 & 36.96 & 29.24 & 26.25 & 25.45 & 37.64 \\
        $\infty$ & 51.62 & 49.36 & 45.85 & \textbf{37.09} & 29.12 & 26.06 & 25.03 & 37.73 \\
        \bottomrule
    \end{tabular}
    \caption{Ablation study on the number of $\pi$ for training label shift adapter using CIFAR-100-C. Num. denotes the number of $\pi$ for training the adapter. F, U, and B indicate forward, uniform, and backward distributions, respectively. We chose three label distributions.}
    \label{tab:ablation_study}
\end{table}

\noindent\textbf{Ablation Study on $\pi$.}
As described in the main manuscript, we sampled three kinds of label distributions for $\pi$ during training label shift adapter.
Regarding the effect of sampling different numbers of $\pi$, Table~\ref{tab:ablation_study} indicates that such variations have negligible impact on performance.
Specifically, in this experiment, we interpolate three distributions (\textit{i.e.}, $\pi_s$, $u$, $\bar{\pi}_s$) and $\tau$ to train the label shift adapter when different numbers of $\pi$ are utilized.

\section*{C. Additional Experiments}

\begin{table}[t!]
    \centering
    \small
    \begin{tabular}{c|ccc}
        \toprule
        & DELTA & ISFDA & TENT+Ours \\
        \midrule
        VISDA-C & 50.10 & 61.02 & \textbf{72.97} \\
        \bottomrule
    \end{tabular}
    \caption{Comparison with additional baselines in test-time adaptation setting.}
    \label{Table:comparison_with_baselines}
\end{table}

\noindent\textbf{Comparison with Baselines Related to Label Shifts.}
We've compared two baselines in Table~\ref{Table:experiment-simple-baseline}, which have the capability of handling label shifts.
We compare additional baselines, DELTA~\cite{zhao2023delta} and ISFDA~\cite{li2021imbalanced}, which address covariate and label shifts simultaneously. 
Although ISFDA requires several epochs for adapting the source models, we conduct the experiments in the test-time adaptation setting for a fair comparison.
Table~\ref{Table:comparison_with_baselines} demonstrates that our method is superior to baselines significantly in the VISDA-C dataset.
ISFDA, a domain adaptation model, exhibits limitations in its suitability for online learning during inference.
Since DELTA only focuses on class imbalances in the target domain, it lacks the ability to handle imbalances in the source domain.
In contrast, our method successfully addresses the imbalance in both source and target domains in the test-time adaptation setting.

\begin{table}[t!]
    \centering
    \scriptsize
    \setlength{\tabcolsep}{3pt}
    \begin{tabular}{@{}c@{}|@{}c@{}|cccccccc@{}}
        \toprule
         & Method & F50 & F25 & F10 & U & B10 & B25 & B50 & Avg.\\
        \midrule
        \multirowcell{3}{CIFAR10} & SAR+GN & 57.22 & 57.20 & 57.07 & 57.12 & 61.84 & 63.06 & 64.37 & 59.70 \\
         & SAR+BN & 78.63 & 76.28 & 71.82 & 53.28 & 34.99 & 28.60 & 25.18 & 52.68 \\
         & Ours+IABN & \textbf{80.58} & \textbf{78.62} & \textbf{75.26} & \textbf{63.34} & \textbf{68.54} & \textbf{70.07} & \textbf{71.64} & \textbf{72.58} \\
        \midrule
        \multirowcell{3}{CIFAR100} & SAR+GN & 9.09 & 9.59 & 10.23 & 14.05 & 18.93 & 20.46 & 21.70 & 14.86 \\
         & SAR+BN & 49.44 & 47.04 & 43.39 & 32.18 & 20.22 & 16.24 & 13.96 & 31.78 \\
         & Ours+IABN & \textbf{52.06} & \textbf{49.71} & \textbf{46.03} & \textbf{36.84} & \textbf{29.29} & \textbf{26.33} & \textbf{25.50} & \textbf{37.97} \\
        \bottomrule
    \end{tabular}
    \caption{Comparison with SAR using CIFAR-10-C and CIFAR-100-C in TTA setting. F, U, and B denote forward, uniform, and backward, respectively. GN and BN indicate group and batch normalization, respectively.}
    \vspace{-0.4cm}
    \label{Table:comparison_with_sar}
\end{table}

\noindent\textbf{Comparison with Recent TTA Baseline.}
We compare recent test-time adaptation baseline, sharpness-aware and reliable entropy minimization (SAR)~\cite{niu2023towards}.
SAR proposes an optimizer and analyzes normalization layers to resolve imbalances in the target domain.
However, it is important to note that our work addresses imbalances in both the source and target domains.
Table~\ref{Table:comparison_with_sar} demonstrates that our method outperforms both SAR+GN and SAR+BN significantly.
Moreover, it is a viable option to integrate our method with SAR method.

\section*{D. Domain-wise Results}

Table~\ref{Table:experiment-cifar10-domain}, \ref{Table:experiment-cifar100-domain}, \ref{Table:experiment-imagenet-domain} show the average classification accuracy on CIFAR-10-C, CIFAR-100-C, and ImageNet-C, shown per domain.
To compute the accuracy of each domain, we calculate the average performances of Forward50, Forward25, Forward10, Uniform, Backward10, Backward25, and Backward50, as described in the main manuscript.
These results demonstrate that our method consistently enhances performance across various domains.



\begin{table*}[t!]
    \small
    \centering
    \resizebox{\linewidth}{!}{
    \begin{tabular}{@{}l@{\quad}cccccccccccccccc@{}}
        \toprule
        \textbf{Method} & \rotatebox{70}{Gaussian} & \rotatebox{70}{Shot} & \rotatebox{70}{Impulse} & \rotatebox{70}{Defocus} & \rotatebox{70}{Glass} & \rotatebox{70}{Motion} & \rotatebox{70}{Zoom} & \rotatebox{70}{Snow} & \rotatebox{70}{Frost} & \rotatebox{70}{Fog} & \rotatebox{70}{Brightness} & \rotatebox{70}{Contrast} & \rotatebox{70}{Elastic} & \rotatebox{70}{Pixelate} & \rotatebox{70}{JPEG} & \textbf{Avg} \\
        \midrule
        Source & 23.06 & 26.94 & 19.59 & 46.89 & 41.06 & 45.63 & 49.42 & 58.30 & 45.31 & 45.72 & 69.85 & 21.23 & 57.60 & 60.14 & 65.45 & 45.08 \\
        \midrule
        BN Stats & 49.18 & 50.52 & 46.29 & 58.11 & 45.05 & 56.14 & 55.96 & 52.32 & 50.65 & 53.60 & 59.11 & 54.93 & 51.63 & 54.15 & 52.12 & 52.65 \\
        ONDA & 50.70 & 51.66 & 47.68 & 59.80 & 46.49 & 57.45 & 57.78 & 53.81 & 52.36 & 55.14 & 61.52 & 54.76 & 53.60 & 56.49 & 54.19 & 54.23 \\
        PseudoLabel & 46.87 & 48.76 & 44.47 & 55.96 & 43.87 & 53.89 & 53.46 & 49.96 & 48.70 & 50.91 & 56.34 & 52.57 & 49.29 & 51.88 & 50.16 & 50.47 \\
        LAME & 17.97 & 22.75 & 15.43 & 44.74 & 40.36 & 42.40 & 47.08 & 61.30 & 48.62 & 45.20 & 67.84 & 20.33 & 55.40 & 61.36 & 64.59 & 43.69 \\
        CoTTA & 51.69 & 53.11 & 50.31 & 55.80 & 47.28 & 54.31 & 54.88 & 52.60 & 52.18 & 52.81 & 58.30 & 49.66 & 52.12 & 55.32 & 54.39 & 52.98 \\
        NOTE & 54.48 & 56.22 & 53.24 & 68.20 & 48.64 & 64.87 & 65.09 & 65.56 & 64.43 & 64.08 & 73.33 & 67.59 & 60.24 & 66.95 & 67.26 & 62.68 \\
        \midrule
        TENT & 46.41 & 48.29 & 43.38 & 53.82 & 42.42 & 52.22 & 51.57 & 48.92 & 47.51 & 49.81 & 55.06 & 50.62 & 47.90 & 50.58 & 49.20 & 49.18 \\
        \rowcolor{gray!10} + Ours & 51.66 & 53.55 & 48.66 & 60.74 & 46.94 & 58.77 & 57.37 & 55.04 & 53.48 & 56.00 & 62.05 & 57.68 & 54.11 & 57.48 & 55.07 & 55.24 \\
        \midrule
        IABN & 54.77 & 56.48 & 53.25 & 68.24 & 48.39 & 64.53 & 64.89 & 65.63 & 64.44 & 64.60 & 73.79 & 67.24 & 60.32 & 67.81 & 67.17 & 62.77 \\
        \rowcolor{gray!10} + Ours & \textbf{68.72} & \textbf{69.54} & \textbf{64.94} & \textbf{77.48} & \textbf{61.47} & \textbf{76.24} & \textbf{75.52} & \textbf{72.06} & \textbf{72.83} & \textbf{74.53} & \textbf{79.69} & \textbf{79.08} & \textbf{69.66} & \textbf{74.25} & \textbf{72.69} & \textbf{72.58} \\
        \bottomrule
    \end{tabular}}
    \caption{\textbf{Domain-wise results on CIFAR-10-C.}}
    \label{Table:experiment-cifar10-domain}
    
\end{table*}




\begin{table*}[t!]
    \small
    \centering
    \resizebox{\linewidth}{!}{
    \begin{tabular}{@{}l@{\quad}cccccccccccccccc@{}}
        \toprule
        \textbf{Method} & \rotatebox{70}{Gaussian} & \rotatebox{70}{Shot} & \rotatebox{70}{Impulse} & \rotatebox{70}{Defocus} & \rotatebox{70}{Glass} & \rotatebox{70}{Motion} & \rotatebox{70}{Zoom} & \rotatebox{70}{Snow} & \rotatebox{70}{Frost} & \rotatebox{70}{Fog} & \rotatebox{70}{Brightness} & \rotatebox{70}{Contrast} & \rotatebox{70}{Elastic} & \rotatebox{70}{Pixelate} & \rotatebox{70}{JPEG} & \textbf{Avg} \\
        \midrule
        Source & 14.42 & 16.45 & 9.08 & 22.55 & 23.14 & 25.65 & 25.82 & 27.37 & 22.98 & 18.57 & 34.51 & 5.84 & 33.03 & 17.47 & 36.19 & 22.21 \\
        \midrule
        BN Stats & 27.59 & 28.51 & 26.20 & 36.89 & 28.61 & 34.53 & 36.44 & 29.40 & 28.81 & 28.98 & 36.66 & 29.22 & 33.37 & 33.78 & 32.18 & 31.41 \\
        ONDA & 27.53 & 28.77 & 26.17 & 36.85 & 29.07 & 34.37 & 36.91 & 29.71 & 29.09 & 29.38 & 36.97 & 27.77 & 33.89 & 34.00 & 33.00 & 31.57 \\
        PseudoLabel & 27.44 & 28.54 & 25.59 & 34.92 & 27.78 & 32.83 & 34.56 & 28.58 & 27.40 & 28.58 & 34.96 & 26.01 & 31.58 & 32.86 & 31.38 & 30.20 \\
        LAME & 13.41 & 15.73 & 7.66 & 20.93 & 22.30 & 24.79 & 24.63 & 27.21 & 22.39 & 17.07 & 33.62 & 4.48 & 32.41 & 15.62 & 35.66 & 21.19 \\
        CoTTA & 30.01 & 30.85 & 28.45 & 34.77 & 30.64 & 34.04 & 35.64 & 30.92 & 30.10 & 28.65 & 36.09 & 24.30 & 33.54 & 35.58 & 34.00 & 31.84 \\
        NOTE & 24.17 & 25.64 & 18.62 & 35.73 & 28.08 & 36.89 & 37.48 & 34.91 & 33.95 & 29.13 & 41.47 & 33.93 & 36.29 & 32.01 & 36.13 & 32.30 \\
        \midrule
        TENT & 27.50 & 28.49 & 25.28 & 33.98 & 26.89 & 32.12 & 33.36 & 28.00 & 26.79 & 27.78 & 34.05 & 25.20 & 31.01 & 32.25 & 30.53 & 29.55 \\
        \rowcolor{gray!10} + Ours & 30.95 & 32.21 & \textbf{28.77} & 38.60 & 30.58 & 36.06 & 38.24 & 32.30 & 30.74 & 31.52 & 38.34 & 30.36 & 34.75 & 36.19 & 34.86 & 33.63 \\
        \midrule
        IABN & 24.54 & 25.79 & 18.92 & 35.50 & 28.00 & 36.80 & 37.58 & 34.97 & 34.20 & 29.02 & 41.39 & 33.99 & 36.17 & 32.09 & 36.29 & 32.35 \\
        \rowcolor{gray!10} + Ours & \textbf{33.65} & \textbf{34.37} & 28.17 & \textbf{41.86} & \textbf{33.62} & \textbf{41.08} & \textbf{41.79} & \textbf{37.82} & \textbf{38.36} & \textbf{34.77} & \textbf{43.51} & \textbf{42.54} & \textbf{39.18} & \textbf{39.98} & \textbf{38.76} & \textbf{37.97} \\
        \bottomrule
    \end{tabular}}
    \caption{\textbf{Domain-wise results on CIFAR-100-C.}}
    \label{Table:experiment-cifar100-domain}
\end{table*}




\begin{table*}[t!]
    \small
    \centering
    \resizebox{\linewidth}{!}{
    \begin{tabular}{@{}l@{\quad}cccccccccccccccc@{}}
        \toprule
        \textbf{Method} & \rotatebox{70}{Gaussian} & \rotatebox{70}{Shot} & \rotatebox{70}{Impulse} & \rotatebox{70}{Defocus} & \rotatebox{70}{Glass} & \rotatebox{70}{Motion} & \rotatebox{70}{Zoom} & \rotatebox{70}{Snow} & \rotatebox{70}{Frost} & \rotatebox{70}{Fog} & \rotatebox{70}{Brightness} & \rotatebox{70}{Contrast} & \rotatebox{70}{Elastic} & \rotatebox{70}{Pixelate} & \rotatebox{70}{JPEG} & \textbf{Avg} \\
        \midrule
        Source & 5.72 & 5.88 & 4.82 & 15.48 & 11.17 & 17.98 & 18.73 & 21.25 & 15.21 & 28.04 & 44.67 & 28.63 & 39.16 & 29.56 & 34.77 & 21.40 \\
        \midrule
        BN Stats & 29.27 & 28.90 & 25.46 & 25.59 & 22.94 & 33.14 & 32.91 & 28.37 & 25.15 & 40.06 & 45.27 & 40.19 & 43.59 & 41.84 & 39.74 & 33.49 \\
        ONDA & 29.15 & 28.88 & 25.33 & 25.49 & 22.74 & 32.73 & 32.96 & 28.18 & 25.04 & 40.12 & 45.46 & 39.69 & 43.60 & 42.02 & 39.72 & 33.41 \\
        PseudoLabel & 31.25 & 30.93 & 29.00 & 27.72 & 25.80 & 34.36 & 33.64 & 30.20 & 25.39 & 40.33 & 44.09 & 39.56 & 42.78 & 41.43 & 39.65 & 34.41 \\
        LAME & 5.56 & 5.73 & 4.67 & 15.33 & 11.03 & 17.92 & 18.67 & 21.19 & 15.16 & 28.01 & 44.64 & 28.61 & 39.12 & 29.49 & 34.75 & 21.33 \\
        CoTTA & 30.93 & 30.36 & 27.47 & 27.28 & 24.95 & 34.42 & 33.56 & 30.01 & 26.44 & 40.62 & 44.79 & 40.58 & 43.32 & 41.83 & 40.05 & 34.44 \\
        NOTE & 31.34 & 30.83 & 29.26 & 27.39 & 24.66 & 35.67 & 32.70 & 33.34 & 28.33 & 39.52 & 46.55 & 44.97 & 43.67 & 41.38 & 41.59 & 35.41 \\
        \midrule
        TENT & 29.29 & 28.94 & 25.84 & 25.68 & 22.94 & 32.11 & 32.37 & 27.17 & 23.50 & 39.43 & 43.94 & 38.18 & 42.30 & 40.66 & 39.02 & 32.76 \\
        \rowcolor{gray!10} + Ours & 32.38 & 32.39 & 28.77 & 29.07 & 26.21 & 35.70 & \textbf{35.81} & 31.44 & 27.86 & \textbf{43.14} & 48.56 & 43.00 & 46.59 & 44.92 & 43.59 & 36.63 \\
        \midrule
        IABN & 31.47 & 30.86 & 29.28 & 27.37 & 24.66 & 35.69 & 32.77 & 33.35 & 28.37 & 39.54 & 46.65 & 45.03 & 43.72 & 41.40 & 41.60 & 35.45 \\
        \rowcolor{gray!10} + Ours & \textbf{34.28} & \textbf{34.00} & \textbf{32.18} & \textbf{30.25} & \textbf{27.14} & \textbf{38.92} & 35.53 & \textbf{36.48} & \textbf{31.58} & 42.50 & \textbf{49.95} & \textbf{48.35} & \textbf{46.93} & \textbf{44.95} & \textbf{44.87} & \textbf{38.53} \\
        \bottomrule
    \end{tabular}}
    \caption{\textbf{Domain-wise results on ImageNet-C.}}
    \label{Table:experiment-imagenet-domain}
\end{table*}




\end{document}